\documentclass[journal]{IEEEtran}

\IEEEoverridecommandlockouts                              

\let\labelindent\relax
\usepackage[table,xcdraw]{xcolor}
\usepackage{subfig}
\usepackage{xspace}
\usepackage{graphicx,url}
\usepackage[font=small]{caption}
\usepackage{times,amsmath,epsfig,amssymb,graphicx,amsfonts}
\usepackage{empheq}
\usepackage{psfrag}
\usepackage{fge}
\usepackage{amsmath}
\usepackage{setspace}
\usepackage{color}
\usepackage{amsfonts}   
\usepackage{amsthm}
\usepackage{amssymb}    
\usepackage{mathrsfs}
\usepackage{setspace}
\usepackage{dblfloatfix}
\usepackage{tikz}
\usepackage{tkz-tab}

\usepackage{pgfplotstable}
\usepackage{pgfplots}
\usepackage{algorithm}
\usepackage{algcompatible}
\usepackage{lipsum}
\usepackage{color}
\usepackage{comment}
\usepackage{mathrsfs}
\usepackage{epstopdf}
\usepackage{enumerate}
\usepackage{enumitem}
\usepackage{url}
\usepackage{cite}
\usepackage{tabularx}
\usepackage[pass]{geometry}

\usepackage{multirow}
\usepackage[colorlinks,bookmarksopen,bookmarksnumbered,citecolor=red,urlcolor=red,hypertexnames=false]{hyperref}

\newtheorem{problem}{Problem}

\newcommand{\bdmath}{\begin{dmath}}
\newcommand{\edmath}{\end{dmath}}
\newcommand{\beq}{\begin{equation}}
\newcommand{\eeq}{\end{equation}}
\newcommand{\bdm}{\begin{displaymath}}
\newcommand{\edm}{\end{displaymath}}
\newcommand{\bea}{\begin{eqnarray}}
\newcommand{\eea}{\end{eqnarray}}
\newcommand{\beal}{\beq \begin{array}{lll}}
\newcommand{\eeal}{\end{array} \eeq}
\newcommand{\beas}{\begin{eqnarray*}}
\newcommand{\eeas}{\end{eqnarray*}}
\newcommand{\ba}{\begin{array}}
\newcommand{\ea}{\end{array}}
\newcommand{\bit}{\begin{itemize}}
\newcommand{\eit}{\end{itemize}}
\newcommand{\ben}{\begin{enumerate}}
\newcommand{\een}{\end{enumerate}}

\newcommand{\calA}{{\cal A}}
\newcommand{\calB}{{\cal B}}

\newcommand{\calI}{{\cal I}}

\newcommand{\calL}{{\cal L}}

\newcommand{\calP}{{\cal P}}

\newcommand{\calR}{{\cal R}}
\newcommand{\calS}{{\cal S}}

\newcommand{\calY}{{\cal Y}}
\newcommand{\calU}{{\cal U}}
\newcommand{\calV}{{\cal V}}

\newcommand{\setN}{\textsf{N}}

\newcommand{\hide}[1]{}

\newcommand{\hiddenText}{{\color{gray} hidden text.}}
\newcommand{\hideWithText}[1]{\hiddenText}

\newcommand{\diag}[1]{\mathrm{diag}\left(#1\right)}

\newcommand{\at}[1]{^{(#1)}}

\newcommand{\SEtwo}{\ensuremath{\mathrm{SE}(2)}\xspace}

\newcommand{\scenario}[1]{{\small \sf#1}\xspace}

\newcommand{\przero}{\eqref{preq:active_inf_acq}\xspace}
\newcommand{\prone}{\eqref{preq:resil_active_inf_acq}\xspace}
\newcommand{\prtwo}{\eqref{eq:robust_trajectory_planning}\xspace}

\newcommand{\mycomment}[1]{\Statex{{\color{gray}// #1 //}}}
\newcommand{\mysidecomment}[1]{{\color{gray}// #1.}}
\newcommand{\mycolon}{\;:\;}

\newcommand{\myin}{\;\in\;}
\newcommand{\mysubseteq}{\;\subseteq\;}

\renewcommand{\at}[3]{_{#1, \; #2 : #3}}

\newcommand{\metric}{J}
\newcommand{\optmetric}{\metric^\star}
\newcommand{\attack}{\alpha}

\newcommand{\optval}[2]{\optmetric_{#1,\; t, \; #2}}

\newcommand{\myalg}{{\sf \small RAIN}\xspace}
\newcommand{\mainproblem}{{\small RAIN}}
\newcommand{\trajectoryproblem}{{\small RTP}}
\newcommand{\algrobust}{{\small \sf RTP}\xspace}

\newcommand{\planninghorizon}{{\small \sf T_{\sf{PLAN}}}}
\newcommand{\missionlength}{{\small\sf T_{\sf{TASK}}}}
\newcommand{\replanrate}{{\small\sf T_{\sf{REPLAN}}}}

\newcommand{\domainoptshortRAIN}[1]{{\substack{ u_{#1,t} \myin \calU_{#1,t} \\ t \;=\; 1\mycolon\missionlength}}}
\newcommand{\optmshortRAIN}[1]{\displaystyle\max_{\domainoptshortRAIN{#1}}}
\newcommand{\domainoptshort}[1]{{\substack{ u_{#1,t'} \myin \calU_{#1,t'}\\ t' \;=\; t+1\mycolon t+\planninghorizon}}}
\newcommand{\attackoptshortt}{\displaystyle\min_{\substack{ \calA_{t} \mysubseteq \calV, \;|\calA_t|\;\leq\; \attack\\ t \;=\; 1\mycolon\missionlength}}}
\newcommand{\optmshort}[1]{\displaystyle\max_{\domainoptshort{#1}}}
\newcommand{\domainoptshortSingleRobot}{{\substack{ u_{i,t'} \myin \calU_{i,t'} \\ t' \;= \;t+1\mycolon t+\planninghorizon}}}
\newcommand{\optmshortSingleRobot}{\displaystyle\max_\domainoptshortSingleRobot}

\renewcommand{\SEtwo}{\ensuremath{\mathrm{SE}(2)}\xspace}

\newcommand{\baitset}{\calL}

\newcommand{\targetprocess}{g}

\newcommand{\myParagraph}[1]{{\bf #1.}\xspace}

\newcommand{\validated}[2]{{#2}}
\newcommand{\eg}{\emph{e.g.},\xspace}
\newcommand{\ie}{\emph{i.e.},\xspace}

\newcommand{\elem}{{{v}}}
\newcommand{\function}{{{f}}}

\floatname{algorithm}{Algorithm}

\floatname{algorithm}{Algorithm}

\newtheorem{mydef}{Definition}

\newtheorem{mytheorem}{Theorem}
\newtheorem{mylemma}{Lemma}
\newtheorem{myremark}{Remark}
\newtheorem{mycorollary}{Corollary}
\newtheorem{myproposition}{Proposition}

\newtheorem*{problemno}{Problem}

\DeclareMathOperator{\sinc}{sinc}

\newcounter{ale}

\newenvironment{liste}{\begin{itemize}}{\end{itemize}}
\newcommand{\aliste}{\begin{liste} \setcounter{ale}{1}}
\newcommand{\zliste}{\end{liste}}

\newcommand{\scaleMathLine}[2][1]{\resizebox{#1\linewidth}{!}{$\displaystyle{#2}$}}

\title{Resilient Active Information Acquisition with\\ Teams of Robots}
\author{Brent Schlotfeldt,{$^{1}$}~\IEEEmembership{Student Member,~IEEE,}
Vasileios Tzoumas,{$^{2}$}~\IEEEmembership{Member,~IEEE,}
George J.~Pappas,{$^{1}$}~\IEEEmembership{Fellow,~IEEE}
\thanks{$^{1}$B.~Schlotfeldt and G.~J.~Pappas are with the Department of Electrical and Systems Engineering, University of Pennsylvania, Philadelphia, PA 19104, USA. {\tt\footnotesize \{brentsc, pappasg\}@seas.upenn.edu}}
\thanks{$^{2}$V.~Tzoumas is with the Department of Aerospace Engineering, University of Michigan, Ann Arbor, MI 48109, USA. {\tt\footnotesize vtzoumas@umich.edu}}
\thanks{This work was partially supported by ARL CRA DCIST W911NF-17-2-0181 and the Rockefeller Foundation.
}
}

\begin{document}
\maketitle

\begin{abstract}
Emerging applications of collaborative autonomy, such as \emph{Multi-Target Tracking}, \emph{Unknown Map Exploration}, and \emph{Persistent Surveillance}, require robots plan paths to navigate an environment while maximizing the information collected via on-board sensors.  {In this paper, we consider such information acquisition tasks but in adversarial environments, where attacks may temporarily disable the robots' sensors.}
We propose the first receding horizon algorithm, aiming for robust and adaptive multi-robot planning against any number of attacks, which we call \emph{Resilient Active Information acquisitioN} (\myalg). \myalg calls, in an online fashion, a \emph{Robust Trajectory Planning} (\algrobust) subroutine which plans attack-robust control inputs over a look-ahead planning horizon.  {We quantify \algrobust's performance by bounding its suboptimality.  We base our theoretical analysis on notions of curvature introduced in combinatorial optimization.}
We evaluate \myalg in three information acquisition scenarios: \emph{Multi-Target Tracking}, \emph{Occupancy Grid Mapping}, and \emph{Persistent Surveillance}.  The scenarios are simulated in C++ and a Unity-based simulator. In all simulations, \myalg runs in real-time, and exhibits superior performance against a state-of-the-art baseline information acquisition algorithm, even in the presence of a high number of attacks.  We also demonstrate \myalg's robustness and effectiveness against varying models of attacks (worst-case and random), as well as, varying replanning rates.
\end{abstract}

\begin{IEEEkeywords}
Autonomous Robots; Multi-Agent Systems; Reactive Sensor-Based Mobile Planning; Robotics in Hazardous Fi- elds; Algorithm Design \& Analysis; Combinatorial Mathematics.
\end{IEEEkeywords}

\section{Introduction}\label{sec:Intro}

\begin{figure}
\centering
\includegraphics[width=.97\linewidth]{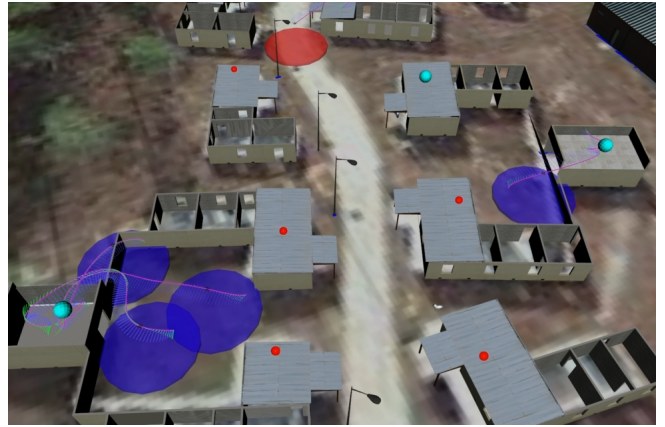}
\caption{\textbf{Persistent Surveillance under Attacks.} Unity simulation environment depicting a $5$-robot team engaging in a \emph{Persistent Surveillance} task for monitoring a set of buildings.  Some robots are under attack. {The attacks can disable the sensing capabilities of the robots, at least temporarily}.  Each blue disc indicates the field of view of a (non-attacked) robot, while each red disc indicates an attacked robot. 
In this adversarial environment, the robots must resiliently plan trajectories to (re-)visit all the building landmarks to continue acquiring information despite the attacks.
}
  \label{fig:sim_ex}
\end{figure}

\emph{Active information acquisition} has a long history in robotics \cite{bajcsy1988active, hero2011sensor}. The  \emph{``active''} characterization captures the idea that robots which move \emph{purposefully} in the environment, \emph{acting} as mobile sensors instead of static, can achieve superior sensing performance. Indeed, active information acquisition has recently been extended to a variety of collaborative (multi-robot) autonomy tasks, such as \emph{Multi-Target Tracking} \cite{atanasov2014information,schlotfeldt2018anytime}, \emph{Exploration and Mapping} (including \emph{Simultaneous Localization and Mapping}) \cite{charrow2014approximate, atanasov2015decentralized,corah2019distributed}, \emph{Monitoring of Hazardous Environments} \cite{atanasov2014information, thakur2018nuclear}, and \emph{Persistent Surveillance} \cite{hilal2013intelligent, smith2011persistent, kumar2004robot}. 

Robotics research has enabled the solution of several classes of information acquisition problems, both for single-robot and multi-robots scenarios, with methods that include search-based \cite{atanasov2014information, schlotfeldt2019maximum}, sampling-based  \cite{hollinger2014sampling, kantaros2019asymptotically}, and gradient-based planning \cite{schwager2017multi}. The methods can also differ on problem parameters, such as the length of the planning horizon ---cf.~\emph{myopic} (or \emph{one-step-ahead}) planning \cite{zhou2019distributed} versus \emph{non-myopic} (or \emph{long-horizon}) planning \cite{atanasov2014information}--- and the type of target process (\eg Gaussian \cite{le2009trajectory}, Gaussian mixture, \cite{charrow2014approximate, tabib2019real},  occupancy grid map \cite{corah2019distributed, charrow2015information}, and non-parametric \cite{hoffmann2009mobile}).

{But most robotics research places no emphasis on robustifying  information acquisition against attacks (or failures) that may temporarily disable the robots' sensors; \eg smoke-cloud attacks that can block temporarily the field of view of multiple sensors.  For example, \cite{saulnier2017resilient} focuses on formation control, instead of information acquisition; \cite{mitra2019resilient} focuses on state estimation against byzantine attacks, \ie attacks that corrupt the robots' sensors with adversarial noise, instead of disabling them; and~\cite{prorok2020robust} focuses on a  trajectory scheduling in transportation networks when travel times can be uncertain, instead on trajectory planning for information acquisition.
An exception is \cite{zhou2019distributed}, which however is limited to multi-target tracking
based on myopic planning, instead of non-myopic. 
}

Related work has also been developed in combinatorial optimization \cite{nemhauser1978analysis, krause2012submodular}, paving the way for robust combinatorial optimization against attacks: \cite{orlin2015robust} proposes algorithms for submodular set function optimization against worst-case attacks, but under the assumption the attacks can remove (disable) only a limited number of elements from the optimized set. Instead, \cite{tzoumas2017resilient} proposes algorithms for optimization against any number of removals.  
And, recently, \cite{tzoumas2018resilient} extended the algorithms to matroid constraints, enabling the application of the algorithms in robotics since multi-robot path planning can be cast as a matroid-constrained optimization problem \cite{corah2019distributed}. 

\myParagraph{Contributions} In this paper, in contrast to the aforementioned works, we extend the \emph{attack-free active information acquisition} to account for attacks against the robots while simultaneously performing non-myopic multi-robot planning. 
We make the following three key contributions.

\myParagraph{1.~Receding Horizon Formulation and Algorithm}  {Section~\ref{sec:prob_form} formalizes the attack-aware active information acquisition problem as a finite horizon control optimization problem named \emph{Resilient Active Information acquisitioN} \prone~---in the acronym, the ``{P}'' stands for ``Problem.'' \prone is a sequential mixed-integer optimization problem: it jointly optimizes the robots' control inputs such that the robots' planned paths are robust against worst-case attacks that may occur at each time step.  The upper-bound to the number of attacks is assumed known and constant.  \prone's information objective function is assumed non-decreasing in the number of robots (a natural assumption, since the more robots the more information one typically can collect).} 

{Section~\ref{sec:alg} proposes \myalg, the first receding horizon algorithm for the problem of resilient active information acquisition \prone. \myalg calls in an online fashion \emph{Robust Trajectory Planning} (\algrobust), a subroutine that plans attack-robust control inputs over a planning horizon.  \algrobust's  planning horizon is typically less than the acquisition problem's finite horizon, for computational reasons.  For the same reason, \algrobust assumes constant attacks. \algrobust is presented in Section~\ref{subsec:alg_description}.
}

\myParagraph{2.~Performance Analysis} 
{Although no performance guarantees exist for the non-linear combinatorial optimization problem \prone, Section~\ref{sec:performance} provides suboptimality bounds on \algrobust's performance, \ie on the algorithm used by \myalg to approximate a solution to \prone \emph{locally}, in a receding horizon fashion.
The theoretical analysis is based on notions of curvature introduced for combinatorial optimization; namely, the notions of \emph{curvature}~\cite{conforti1984curvature} and \emph{total curvature}~\cite{sviridenko2017optimal}.  The notions aim to bound the worst-case complementarity of the robots' planned  paths in their ability to jointly maximize \prone information acquisition objective function.
}

\myParagraph{3.~Experiments} {Section~\ref{sec:simulations_experiments} evaluates \myalg across three multi-robot information acquisition tasks: \emph{Multi-Target Tracking}, \emph{Occupancy Grid Mapping}, and \emph{Persistent Surveillance}.  All evaluations demonstrate the necessity for  attack-resilient planning, via a comparison with a state-of-the-art baseline information acquisition algorithm, namely, \emph{coordinate descent}~\cite{atanasov2015decentralized}. 
Specifically, \myalg runs in real-time and exhibits superior performance in all experiments.  \myalg's effectiveness is accentuated the higher the numbers of attacks is (Section~\ref{subsec:multi-target}). \myalg remains effective even against non-worst-case attacks, specifically, random (Section~\ref{subsec:occupancy}).  Even when high replanning rates are feasible (Section~\ref{subsec:persistent}), in which case coordinate descent can adapt at each time step against the observed attacks, \myalg still exhibits superior performance.
The algorithm is implemented in C++ and a Unity-based simulator.}

\myParagraph{Comparison with Preliminary Results in~\cite{Schlotfeldt18iros-RelientInfGath}}
This paper extends the results in~\cite{Schlotfeldt18iros-RelientInfGath}, by introducing  novel problem formulations, algorithms, and numerical evaluations, as well as, by including all proofs (Appendix), which were omitted from~\cite{Schlotfeldt18iros-RelientInfGath}.  {The receding horizon formulation \prone is novel, generalizing the \prtwo formulation first presented in~\cite{Schlotfeldt18iros-RelientInfGath}.  The algorithm \myalg is also first presented here.
Additionally, the simulation evaluations on \emph{Occupancy Grid Mapping} and \emph{Persistent Surveillance} are new, and have not been previously published.  They also include for the first time
a sensitivity analysis of \myalg against varying models of attacks (worst-case and random), as well as, varying replanning rates. 
}

 \section{Resilient Active Information acquisitioN (\mainproblem) Problem}\label{sec:prob_form}

We present the optimization problem of \emph{Resilient Active Information acquisitioN} (\mainproblem) (Section~\ref{sec:problem_definition}). {To this end, we first formalize the (attack-free) \emph{active information acquisition problem} (Section~\ref{sec:definitions}). We also use the notation:
\begin{itemize}
    \item $\phi_{\calV,\; \tau\mycolon \tau+\tau'}\triangleq \{\phi_{i,t}\}_{i\myin\calV, \; t\myin [\tau, \ldots, \tau+\tau']}$, for any variable of the form $\phi_{i,t}$, where $\calV$ denotes a set of robots (\ie $i\in\calV$ is robot $i$), and $[\tau, \ldots, \tau+\tau']$ denotes a discrete time interval ($\tau\geq 1$, while $\tau'\geq 0$);
    \item $w \sim\setN({\mu},{\Sigma})$ denotes a Gaussian random variable $w$ with mean $\mu$ and covariance $\Sigma$.
\end{itemize}
}

\subsection{Active Information Acquisition in the \emph{Absence} of Attacks} 
\label{sec:definitions}

{\emph{Active information acquisition} is a control input optimization problem over a finite-length time horizon: it looks to jointly optimize the control inputs for a team of mobile robots so that the robots, acting as mobile sensors, maximize the acquired information about a target process.  Evidently, the optimization must account for the (a) robot dynamics, (b) target process, (c) sensor model, (d) robots' communication network, and (e) information acquisition objective function:}

\setcounter{paragraph}{0}
\paragraph{Robot Dynamics}
We assume noise-less, non-linear robot dynamics, adopting the framework introduced in~\cite{atanasov2015decentralized}:
\begin{align}\label{eq:robot}
x_{i,t} = f_i(x_{i,t-1},\;u_{i,t-1}), \quad i \in \calV, \quad t=1,2,\ldots,
\end{align}
where $\calV$ denotes the set of available robots, $x_{i,t} \in \mathbb{R}^{n_{x_{i,t}}}$ denotes the state of robot~$i$ at time $t$,\footnote{$x_{i,t}$ in eq.~\eqref{eq:robot} belongs to an appropriate state space, such as $SE(2)$ or $SE(3)$, depending on the problem instance.} and $u_{i,t} \in \calU_{i,t}$ denotes the control input to robot $i$; $\calU_{i,t}$ denotes the \textit{finite} set of admissible control inputs to the robot. 

\paragraph{Target Process}  
We assume any target process
\begin{align}\label{eq:target}
    y_t = \targetprocess(y_{t-1})+w_t, \quad t=1,2,\ldots,
\end{align}
where $y_t$ denotes the target's state at time $t$, and $w_t$ denotes {gaussian} process noise; we consider $w_t\sim \setN({\mu}_{w_{t}},{\Sigma}_{w_{t}})$.

\paragraph{Sensor Model}
We assume measurements of the form
\begin{align}\label{eq:sensors}
z_{i,t} = h_i(x_{i,t},\; y_t) + v_{i,t}(x_{i,t}), \quad t=1,2,\ldots,
\end{align}
where $z_{i,t}$ denotes the measurement by robot $i$ at time $t$, and $v_{i,t}$ denotes measurement noise; we consider {$v_{i,t} \sim\setN({\mu}_{v_{i,t}}((x_{i,t}),{\Sigma}_{v_{i,t}}((x_{i,t}))$}.  Both the noise and sensor function $h_i$ depend on  $x_{i,t}$, as it naturally is the case for, \eg bearing and range measurements (cf.~Section~\ref{subsec:multi-target}).

\paragraph{{Communication Network among Robots}}
{We assume centralized communication, \ie all robots can communicate with each other at any time.}

\paragraph{Information Acquisition Objective Function} 
The information acquisition objective function captures the acquired information about the target process, as collected by the robots during the task via their measurements. 
In this paper,  in particular, we consider objective functions of the additive form 
\begin{align}\label{eq:metric}
       {J_{\calV,\; 1:\missionlength}}\triangleq {\sum_{t=1}^{\missionlength} J(y_t\; | \; u_{\calV,1:t},\;  z_{\calV,1:t})},
\end{align}
{where  $\missionlength$ denotes the duration of the information acquisition task, and $J(y_t\; | \; u_{\calV,1:t},\;  z_{\calV,1:t})$  is an information metric such as the \emph{conditional entropy}~\cite{atanasov2015decentralized} (also, cf.~Section~\ref{subsec:multi-target}) or the \emph{mutual information}~\cite{corah2019distributed} (also, cf.~Section~\ref{subsec:occupancy}), where
we make explicit only the metric's dependence on $u_{\calV,1:t}$ and $z_{\calV,1:t}$  (and we make implicit the metric's dependence on the initial conditions $y_0$ and $x_{i,0}$, and on the noise parameters, \ie the means and covariances of $w_t$ and $v_{t}$ for $t=1,\ldots,\missionlength$).
}

{\color{black}
\begin{problemno}[(Attack-Free) Active Information Acquisition]
At time $t=0$, find control inputs $u_{\calV, \; 1:\missionlength}$ by solving the optimization problem
\begin{equation}\label{preq:active_inf_acq} 
\optmshortRAIN{\calV} \quad J_{\calV,\; 1:\missionlength}.
\end{equation}
\end{problemno}
}

{
Eq.~\przero captures a control input optimization problem where across a task-length horizon, the control inputs of all robots are jointly optimized to maximize $J_{\calV,\; 1:\missionlength}$. 
}

{
Solving eq.~\przero can be challenging, mainly due to (i) the non-linearity of eqs.~\eqref{eq:robot}-\eqref{eq:sensors}, (ii) the duration of the task, $\missionlength$, which acts as a look-ahead planning horizon (the longer the planning horizon is, the heavier eq.~\przero is in computing an optimal solution), and (iii) that at $t=0$ \emph{no} measurements have been realized yet.
}

{
To overcome the aforementioned challenges, \emph{on-line} solutions to eq.~\przero have been proposed~\cite{atanasov2015decentralized}, similar to the \emph{Receding Horizon Control} solution ---also known as \emph{Model Predictive Control} (MPC)--- for the finite-horizon optimal control problem~\cite[Chapter~12]{borrelli2017predictive}. Specifically, per the receding horizon approach, one aims to solve eq.~\przero sequentially in time, by solving at each $t=1,\ldots,\missionlength$ an easier version of eq.~\przero, but of the same form as eq.~\przero, where (i) the look-ahead horizon $\missionlength$ is replaced by a shorter $\planninghorizon$ ($\planninghorizon\leq \missionlength$), and (ii) eqs.~\eqref{eq:robot}-\eqref{eq:sensors} are replaced by their~li- nearizations given the current $x_{\calV,t}$ and current estimate of $y_t$.
}

\subsection{Active Information Acquisition in the \emph{Presence} of Attacks} 
\label{sec:problem_definition}

{Eq.~\przero may suffer, however, from an additional challenge: the presence of attacks against the robots, which, if left unaccounted, can compromise the effectiveness of any robot plans per eq.~\przero.  In this paper, in particular, we consider the presence of the following type of attacks:}

\setcounter{paragraph}{5}
\paragraph{{Attack Model}} {At each $t$, an attack $\calA_t$ can \emph{remove} at most $\alpha$ robots from the information acquisition task ($\calA_t\subseteq \calV$ and $|\calA_t| \leq \alpha$), in the sense that any removed robot $i$ ($i \in \calA_t$) cannot acquire any measurement $z_{i,t}$.
In selecting the attack, the attacker has perfect knowledge of the state of the system.  The attacker can select the worst-case attack (cf.~Problem~1).
Nevertheless, the attacker  cannot necessarily prevent the robots from moving according to their pre-planned path, nor can cause any communication loss among the robots. 
}

{\color{black}
\begin{problem}[Problem of Resilient Active Information acquisitioN (P-RAIN)] \label{pr:rain}
At time $t=0$, find control inputs $u_{\calV, \; 1:\missionlength}$ by solving the optimization problem
\begin{equation}\label{preq:resil_active_inf_acq} \tag{{\small P-RAIN}}
\optmshortRAIN{\calV} \;\; \attackoptshortt \quad 
       {J_{\calV\setminus \calA_{t},\; 1:\missionlength}}
\end{equation}
\end{problem}
}

{
\prone goes beyond eq.~\przero by accounting for the attacks $\calA_t$ ($t = 1, \ldots, \missionlength$).  This is expressed in \prone with the minimization step, which aims to prepare an optimal solution $u_{\calV,t}$ against any worst-case attack that may happen at $t$.  
}

{\color{black}
\begin{myremark}[Need for \prone]
Reformulating the (attack-free) eq.~\przero as in \prone may seem unnecessary, since we consider that the attacker cannot cause any communication loss among the non-attacked robots (cf.~the \emph{attack model} defined above): indeed, if the non-attacked robots can \emph{instantaneously observe the attacks at each $t$}, and \emph{instantaneously replan at the same moment $t$}, then \prone is unnecessary. However, replanning instantaneously in practice is impossible, due to (i) computationally-induced and algorithmic delays~\cite{atanasov2015decentralized}, as well as (ii) delays induced by the temporal discretization of the robot and target dynamics.  Thus, for the duration replanning is impossible, the plans need to account for attacks.
\end{myremark}
}
\section{Receding Horizon Approximation:\\
{\small \sf RAIN} Algorithm}\label{sec:alg}

{In solving \prone, one has to overcome not only the  challenges involved in eq.~\przero (cf.~Section~\ref{sec:prob_form}) bult also the additional challenge of the worst-case attacks $\calA_t$ (which are unknown a priori).  We develop an on-line approximation procedure for \prone, summarized in Algorithm~\ref{alg:rain}. 
}

{\color{black}
\setcounter{paragraph}{0}
{\myParagraph{Intuitive Description}} \myalg proposes a receding horizon solution to \prone, {that enables on-line reaction to the history of attacks}, and, thus, is resilient, by executing the  steps:

\paragraph{Initialization \emph{(line~\ref{rain:begin-initialize})}} At $t=0$, the acquisition task has not started and no attacks are assumed possible ($\calA_0=\emptyset$).

\paragraph{Receding Horizon Planning \emph{(lines~\ref{alg:rain-begin-for-receding}-\ref{alg:rain-end-for-receding})}} At each $t=1,\ldots, \missionlength$, \myalg executes the receding horizon steps:

\begin{itemize}
    
    \item \emph{Robust Trajectory Planning \emph{ (\algrobust) (lines~\ref{alg:rain-begin-receding}-\ref{alg:rain-end-receding})}:} Given the current estimate $\hat{y}_t$ of the target process, all robots jointly optimize their control inputs by solving Problem~\ref{pr:rtp}, presented next, which is of the same form as \prone but where (i) the look-ahead horizon $\missionlength$ is replaced by a shorter $\planninghorizon$ ($\planninghorizon\leq \missionlength$), and (ii) the attack is considered fixed over the look-ahead horizon: 
    
    {\color{black}
\begin{problem}[Robust Trajectory Planning (\trajectoryproblem)] \label{pr:rtp}
At time~$t$, find attack-robust control inputs $u_{\calV, \; t+1:t+\planninghorizon}$ by solving the optimization problem
\begin{equation}\label{eq:robust_trajectory_planning}
\tag{{\small P-RTP}}
\optmshort{\calV}\;\;\min_{\calA\;\subseteq\; \calV,\; |\calA|\;\leq\; \attack} \quad
\metric_{\calV\setminus\calA,\; t+1:t+\planninghorizon}.
\end{equation}
\end{problem}
}

Both aforementioned (i) and (ii) intend to make \prtwo computationally tractable, so \prtwo can be solved in real time for the purposes of receding horizon planning. Particularly, we assume that the the algorithm we propose for \prtwo, \algrobust, is called in \myalg every $\replanrate$ steps.\footnote{{{\sf \footnotesize RTP}'s pseudo-code is presented in Algorithm~\ref{alg:dec_resil_coord_decent}, and described in more detail Section~\ref{subsec:alg_description}.  We quantify {\sf \footnotesize RTP}'s performance guarantees in Section~\ref{sec:performance}.}}

    {\color{black}
      \begin{myremark}[Role of $\replanrate$]
      $\replanrate$  is chosen so that a receding horizon plan can always be generated in the duration it takes to compute a solution to \prtwo via \algrobust; 
    \eg if one timestep ---real-time interval from any $t$ to $t+1$--- has duration $0.5 s$, and solving \prtwo via \algrobust requires $2 s$, then $\replanrate=4$ steps.  Generally, $\replanrate \geq 1$ steps.
    Factors that influence the required time to solve \prtwo include the size of the robot team, the length of the planning horizon $\planninghorizon$~\cite{atanasov2014information}, the need for linearization of eqs.~\eqref{eq:robot}-\eqref{eq:sensors}, and the number of possible attacks $\alpha$ ---evidently, the latter factor is unique to \prone, in comparison to the attack-free eq.~\przero.
     \end{myremark}
    }

    \item \emph{Control Execution \emph{(lines~\ref{alg:rain-begin-controlupdate-receding}-\ref{alg:rain-end-controlupdate-receding})}:} Each robot $i$ uses their computed $u_{i,t}$ to make the next step in the environment (in the meantime, the real time changes from $t$ to $t+1$ by the completion of the step).
    
    \item \emph{Attack Observation \emph{(line~\ref{alg:rain-attack-observation})}:} \myalg observes the current attack, which affects the robots while they execute $u_{i,t}$.
    
    \item \emph{Measurement Collection \emph{(lines~\ref{alg:rain-begin-sensing}-\ref{alg:rain-end-sensing})}:} The measurements from all non-attacked robots are collected.
    
    \item \emph{Estimate Update \emph{(line~\ref{alg:rain-estimation})}:} Given all received measurements up to the current time, the estimate of ${y}_t$ is updated.
    \item \emph{Time Update \emph{(line~\ref{alg:rain-timeupdate})}:}  \myalg updates the time counter to match it with the real time.
\end{itemize}
}

\begin{algorithm}[t!]
\caption{\mbox{Resilient \!Active \!Information \!acquisitioN  (\myalg).}}
\begin{algorithmic}[1]
\REQUIRE  \myalg receives the inputs:
\begin{itemize}
    \item \textit{Offline}: Duration $\missionlength$ of information acquisition task; look-ahead horizon $\planninghorizon$ for planning trajectories ($\planninghorizon\leq\missionlength$); replanning rate $\replanrate$ ($\replanrate$ $\leq\planninghorizon$); model dynamics $f_{i}$ of each robot $i$'s state $x_{i,t}$, including initial condition $x_{i,0}$ ($i \in \calV$); sensing model $h_i$ of each robot $i$'s sensors, including ${\mu}_{v_{i,t}}$ and ${\Sigma}_{v_{i,t}}$; model dynamics $\targetprocess$ of target process, including initial condition $y_0$, and ${\mu}_{w_t}$ and covariance ${\Sigma}_{w_{t}}$; objective function $\metric$; number of attacks $\attack$.
    \item \textit{Online}: At each $t=1,\ldots, \missionlength$, observed (i) attack $\calA_t$ (\ie robot removal $\calA_t \subseteq \calV$), and (ii) measurements $z_{i,t}$ from each non-attacked robot $i\in\calV\setminus\calA_t$.
\end{itemize}
\ENSURE At each $t=1,\ldots, \missionlength$, estimate $\hat{y}_t$ of $y_t$.
\mycomment{Initialize}
\STATE{$t=0$;\quad $\hat{y}_t=y_0$;\quad $\calA_t=\emptyset$;\quad $z_{t} = \emptyset$;}\label{rain:begin-initialize}
\mycomment{Execute {resilient} active information acquisition task}
\WHILE{$t< \missionlength$}\label{alg:rain-begin-for-receding}
\Statex{{\color{gray} \quad //(Re)plan robust trajectories for all robots //}}
\IF{$t\!\mod \replanrate = 0$} \label{alg:rain-begin-receding}
\STATE{$\calI_t = \{t,\{f_i,x_{i,t},h_{i},{\mu}_{w_{t}}, {\Sigma}_{w_{t}}, {\mu}_{v_{i,t}}, {\Sigma}_{v_{i,t}}\}_{i\in \calV},\targetprocess, \hat{y}_t,$}
\STATE{\hspace{9mm}$ z_{t}, \planninghorizon, \attack\}$;} \mysidecomment{Denote by $\calI_t$ the information needed by the \algrobust algorithm, called in the next line}
\STATE{${u\at{\calV}{t+1}{t+\planninghorizon}} = \algrobust(\calI_{{0:t}})$;} \mysidecomment{Plan robust trajectories for all robots with look-ahead planning horizon $\planninghorizon$}
\ENDIF\label{alg:rain-end-receding}
\Statex{{\color{gray}\quad // Execute current step of trajectory computed by \algrobust //}}
\FOR{\textbf{all} $i \in \calV$} \label{alg:rain-begin-controlupdate-receding}
\STATE{$x_{i,t+1} = f_i(x_{i, t}, u_{i, t})$;}
\ENDFOR\label{alg:rain-end-controlupdate-receding}
\STATE{Observe $\calA_{t+1}$;}\label{alg:rain-attack-observation} \mysidecomment{Determined by environment/attacker}\! \label{alg:rain-attack}
\Statex{{\color{gray} \quad // Integrate measurements from non-attacked robots //}}
\FOR{\textbf{all} $i \in \calV \setminus \calA_{t+1}$}\label{alg:rain-begin-sensing}
\STATE{Receive measurement $z_{i,t+1}$;} \mysidecomment{Only measurements from non-attacked robots are received}
\ENDFOR\label{alg:rain-end-sensing}
\STATE{\textbf{update}  \label{alg:rain-estimation}
Estimate $\hat{y}_{t+1}$ of $y_{t+1}$ given $z_{1\mycolon t+1}$; \mysidecomment{$z_{1\mycolon t}$ collects all available measurements up to the time $t$, \ie $z_{1\;:\;t}\triangleq\{z_{i,\tau}: i\in\calV\setminus \calA_\tau, \tau=1,\ldots, t\}$}}
\STATE{$t=t+1$;} \mysidecomment{Time update} \label{alg:rain-timeupdate}
\ENDWHILE\label{alg:rain-end-for-receding}
\end{algorithmic} \label{alg:rain}
\end{algorithm}
\section{Robust Trajectory Planning (\algrobust) Algorithm}\label{subsec:alg_description}

{We present \algrobust, which is used as a subroutine in \myalg, in a receding horizon fashion (cf.~Section~\ref{sec:alg}).  \algrobust's pseudo-code is presented in Algorithm~\ref{alg:dec_resil_coord_decent}. \algrobust's performance is quantified in Section~\ref{sec:performance}. We next give an intuitive description of \algrobust. 
}

{\myParagraph{Intuitive Description}  \algrobust's goal is to maximize \prtwo's objective function $\metric_{\calV\setminus\calA,\; t+1:t+\planninghorizon}$, despite a worst-case attack $\calA$ that removes up to $\alpha$ robots from $\calV$.  In~this~context, \algrobust aims to fulfill \prtwo's goal with a two-step process, where (i) \algrobust partitions robots into two sets (the set of robots $\calL$, and the set of robots $\calV\setminus \calL$; cf.~\algrobust's lines~\ref{line1:step_1}-\ref{line1:step_2}), and, then, (ii) \algrobust appropriately selects the robots' control inputs in each of the two sets (cf.~\algrobust's lines~\ref{alg:rtp-begin-second-for}-\ref{line1:step_4}).}  In~particular, \algrobust picks $\calL$ aiming to guess the worst-case removal of $\alpha$ robots  from $\calV$, \ie to guess the optimal solution to the minimization step in \prtwo.  Thus, intuitively, $\calL$ is aimed to act as a ``bait'' to the attacker.  Since guessing the optimal ``bait'' is, in general, intractable~\cite{Feige:1998:TLN:285055.285059}, \algrobust aims to approximate it by letting $\calL$ be the set of $\attack$ robots with the $\attack$ largest marginal contributions to $\metric_{\cdot,\; t+1:t+\planninghorizon}$ (\algrobust's lines~\ref{alg:rtp-begin-second-for}-\ref{alg:rtp-end-second-for}).  Then, \algrobust assumes the robots in $\calL$ are non-existent, and plans the control inputs for the remaining robots (\algrobust's line~\ref{line1:step_4}).

\begin{algorithm}[t!]
\label{alg:rtp}
\caption{Robust Trajectory Planning (\algrobust).}
\begin{algorithmic}[1]
\REQUIRE  
Look-ahead horizon $\planninghorizon$ for planning trajectories; current time $t$; set of robots $\calV$; model dynamics $f_{i}$ of each robot $i$'s state, including current state $x_{i,t}$; sensing model $h_i$ of each robot $i$'s sensors, including ${\mu}_{v_{i,t}}$ and ${\Sigma}_{v_{i,t}}$; model dynamics $\targetprocess$ of target process $y_t$, including current estimate $\hat{y}_t$, and ${\mu}_{w_t}$ and covariance ${\Sigma}_{w_{t}}$; objective function $\metric$; {measurement history $z_{1:t}$}; number of attacks~$\attack$.
\ENSURE Control inputs $u_{i,t'}$, for all robots $i\in\calV$ and all times $t'=t+1,\ldots,t+\planninghorizon$.
\mycomment{Step 1: Generate \emph{bait} robot set (to approximate a worst-case attack assumed constant $\forall t'\in [t+1, t+\planninghorizon]$)}
\FOR{\textbf{all} $i \in \calV$} \mysidecomment{Compute the value of~\eqref{eq:robust_trajectory_planning} assuming (i) only robot $i$ exists and (ii) no attacks will happen}\label{line1:step_1}
\State{{$\optval{\{i\}}{0}\triangleq \optmshortSingleRobot \metric_{\{i\},\;t+1:t+\planninghorizon}$};}\label{line1:opt_1}
\ENDFOR\label{line1:end-step_1}
\STATE{Find a subset $\baitset$ of $\attack$ robots such that ${\optval{\{i\}}{0}}\geq{\optval{\{j\}}{0}}$ for all $i\in \calL$ and $j \in \calV\setminus\calL$;} \mysidecomment{$\baitset$ is the \emph{bait} robot set ($\baitset\subseteq\calV$ and $|\baitset|=\attack$)}\label{line1:step_2}
\FOR{\textbf{all} $i \in \baitset$}\label{alg:rtp-begin-second-for} \mysidecomment{Assign to each robot $i\in \baitset$ the trajectory that achieves {$\optval{\{i\}}{0}$}}
\State{${
u\at{i}{t+1}{t+\planninghorizon} = \arg \optmshortSingleRobot \metric_{\{i\},\;t+1:t+\planninghorizon}}
$;}\label{line1:step_3}
\ENDFOR\label{alg:rtp-end-second-for}
\mycomment{Step 2: Remaining robots, $\calV\setminus\baitset$, plan assuming (i) only robots in $\calV\setminus\baitset$ exists and (ii) no attacks will happen}
\State{${
u\at{\calV\setminus\baitset}{t+1}{t+\planninghorizon} = \arg\! \optmshort{\calV\setminus\baitset}\! \metric_{\calV\setminus\baitset,\;t+1:t+\planninghorizon}
}$.
}\label{line1:step_4}
\end{algorithmic} \label{alg:dec_resil_coord_decent}
\end{algorithm}
\section{Performance Guarantees of {\small \sf RTP}
}\label{sec:performance}

{Performance guarantees are unknown for \myalg, and, correspondingly, \prone (\prone is a mixed-integer, sequential control optimization problem, with limited a priori information on the measurements and attacks that are going to occur during the task-length, look-ahead time horizon).  Nevertheless, in this section we quantify \algrobust's performance, which is used by \myalg in a receding horizon fashion to approximate a solution to \prone locally (cf.~\myalg's lines~\ref{alg:rain-begin-for-receding}-\ref{alg:rain-end-for-receding}), by picking sequentially in time control inputs given (i) a shorter, computationally feasible look-ahead time horizon (cf.~Section~\ref{sec:alg}), and (ii) the history of the so far observed measurements and attacks.  
}

Particularly, in this section we bound \algrobust's approximation performance and running time.  {We use properties of the objective function $\metric_{\calV\setminus \calA,\; t+1:t+\planninghorizon}$ in \prtwo as a function of the set of robots}; namely, the following notions of curvature.

\subsection{Curvature and Total Curvature}\label{sec:total_curvature}
 
We present the notions of \emph{curvature} and \emph{total curvature} for set functions.  We start with the notions of \emph{modularity}, and of \textit{non-decreasing} and \textit{submodular} set functions.

\begin{mydef}[Modularity~{\cite{nemhauser78analysis}}]\label{def:modular}
{Consider a \validated{finite ground}{finite (discrete)} set~$\mathcal{V}$. A set function $h:2^\calV\mapsto \mathbb{R}$ is modular if and only if $h(\calA)=\sum_{\elem\in \calA}h(\elem)$, for any $\calA\subseteq \calV$.}
\end{mydef}

{Hence, if $h$ is modular, then $\calV$'s elements complement each other through $h$. Specifically, Definition~\ref{def:modular} implies $h(\{\elem\}\cup\calA)-f(\calA)= h(\elem)$, for any $\calA\subseteq\calV$ and $\elem\in \calV\setminus\calA$.}

\begin{mydef}[Non-decreasing set function~{\cite{nemhauser78analysis}}]\label{def:mon}
Consider a \validated{finite ground}{finite (discrete)} set~$\mathcal{V}$. $h:2^\calV\mapsto \mathbb{R}$ is 
\emph{non-decreasing} if $h(\validated{\mathcal{A}'}{\mathcal{B}})\geq h(\mathcal{A})$ for all $\mathcal{A}\subseteq \calB$.
\end{mydef}

\begin{mydef}[Submodularity~{\cite[Proposition 2.1]{nemhauser78analysis}}]\label{def:sub}
Consider a finite (discrete) set $\calV$.  $h:2^\calV\mapsto \mathbb{R}$ is \emph{submodular} if 
$h(\mathcal{A}\cup \{\elem\})\!-\!h(\mathcal{A})\geq h(\validated{\mathcal{A}'}{\mathcal{B}}\cup \{\elem\})\!-\!h(\validated{\mathcal{A}'}{\mathcal{B}})$ for all $\mathcal{A}\subseteq \validated{\mathcal{A}'}{\mathcal{B}}$ and $\elem\in \calV$.
\end{mydef}
{Therefore, $h$ is submodular if and only if  the return $h(\calA\cup \{\elem\})-h(\calA)$ diminishes as $\calA$ grows, for any $\elem$. 
If $h$ is submodular, then $\calV$'s elements substitute each other, in contrast to $h$ being modular. Particularly, consider $h$ to be non-negative (without loss of generality): then, Definition~\ref{def:sub} implies $h(\{\elem\}\cup\calA)-h(\calA)\leq f(\elem)$. Thereby, $\elem$'s contribution to $f(\{\elem\}\cup\calA)$'s~value is diminished in the presence of $\calA$.}

\begin{mydef}[Curvature~\cite{conforti1984curvature}]\label{def:curvature}
Consider a finite (discrete) $\mathcal{V}$, and a non-decreasing submodular $h:2^\mathcal{V}\mapsto\mathbb{R}$ such that $h(\elem)\neq 0$ for any $\elem \in \mathcal{V}$, without loss of generality.  Then, $h$'s \emph{curvature} is defined~as 
\begin{equation}\label{eq:curvature}
\kappa_h\triangleq 1-\min_{\elem\in\mathcal{V}}\frac{h(\mathcal{V})-h(\mathcal{V}\setminus\{\elem\})}{h(\elem)}.
\end{equation}
\end{mydef}

{Definition~\ref{def:curvature} implies $\kappa_h \in [0,1]$.   If $\kappa_h=0$, then  $h(\calV)-h(\calV\setminus\{v\})=h(v)$, for all $v\in\calV$, \ie $h$ is modular. Instead, if $\kappa_h=1$, then there exist $v\in\calV$ such that $h(\calV)=h(\calV\setminus\{v\})$, that is, $v$~has no contribution to $h(\calV)$ in the presence of $\calV\setminus\{v\}$.  Overall, $\kappa_h$ represents a measure of how much $\calV$'s elements complement (and substitute) each other.}

\begin{mydef}[Total curvature~{\cite[Section~8]{sviridenko2017optimal}}]\label{def:total_curvature}
Consider a finite (discrete) set $\mathcal{V}$ and a monotone $h:2^\mathcal{V}\mapsto\mathbb{R}$.  Then, $h$'s \emph{total curvature} is quantified as
\begin{equation}\label{eq:total_curvature}
c_h\triangleq 1-\min_{v\in\mathcal{V}}\min_{\mathcal{A}, \mathcal{B}\subseteq \mathcal{V}\setminus \{v\}}\frac{h(\{v\}\cup\mathcal{A})-h(\mathcal{A})}{h(\{v\}\cup\mathcal{B})-h(\mathcal{B})}.
\end{equation}
\end{mydef}

{Definition~\ref{def:total_curvature} implies $c_h\in [0,1]$, similarly to Definition~\ref{def:total_curvature} for $\kappa_h$. When $h$ is  submodular, then $c_h=\kappa_h$. 
Generally, if  $c_f=0$, then $h$ is modular, while if $c_h=1$, then eq.~\eqref{eq:total_curvature} implies Definition~\ref{def:total_curvature}'s assumption that $h$ is non-decreasing.}

\subsection{Performance Analysis for \algrobust}\label{subsec:perf_decentralized_resil_alg}

We quantify (i) suboptimality bounds on \algrobust's approximation performance, and (ii) upper bounds on the
{running time}
\algrobust requires.  {We use the notation:}
{\begin{itemize}
\item $\optval{\calV}{\attack}$ is the optimal value of \prtwo: 
\begin{equation*}
\hspace{-0mm}\optval{\calV}{\attack}\triangleq\!\optmshort{\calV}\min_{\calA\subseteq \calV,\; |\calA|\;\leq\; \attack} 
\metric_{\calV\setminus\calA,\; t+1:t+\planninghorizon};
\end{equation*}
\item $\calA^\star$ is an optimal removal of $\alpha$ robots from $\calV$ per \prtwo:
\begin{equation*}
\calA^\star\triangleq \arg\min_{\calA\subseteq \calV,\; |\calA|\leq \alpha}\; \metric_{\calV\setminus\calA,\; t+1:t+\planninghorizon}.
\end{equation*}
\end{itemize}
}
\noindent{We also use the definitions:}
{\begin{mydef}[Normalized set function~{\cite{nemhauser78analysis}}]\label{def:norm}
Consider a discrete set~$\mathcal{V}$.  $h:2^\calV\mapsto \mathbb{R}$ is
\emph{normalized} if $h(\emptyset) = 0$.
\end{mydef}}

{\begin{mydef}[Non-negativeness~{\cite{nemhauser78analysis}}]\label{def:mon}
Consider a discrete set~$\mathcal{V}$. $h:2^\calV\mapsto \mathbb{R}$ is 
\emph{non-negative} if $h(\mathcal{A})\geq 0$ for all $\mathcal{A}$.
\end{mydef}
}

\begin{mytheorem}[Performance of \algrobust]\label{th:per_alg_dec_resil_coord_decent}
Consider an instance of \prtwo.  Assume the robots in~$\calV$ can solve optimally the 
{(attack-free) information acquisition problem} in eq.~\eqref{preq:active_inf_acq}.

\begin{itemize}
\item{\emph{Approximation performance:}} \algrobust returns control inputs {$u\at{\calV}{1}{t+\planninghorizon}$} such that (i) if {$\metric_{\cdot, t+1:t+\planninghorizon}:2^\calV\mapsto \mathbb{R}$} is non-decreasing, and, without loss of generality, normalized and non-negative, then
    \begin{equation}\label{ineq:bound_non_sub}
	 \frac{
	 {
	 \metric_{\calV\setminus\calA^\star,\;t+1:t+\planninghorizon}
	 }
	 }{{\optval{\calV}{\attack}}}\geq (1-c_{{\metric_{\cdot, t+1:t+\planninghorizon}}})^2;
	\end{equation}

(ii) If, in addition, {$\metric_{\cdot, t+1:t+\planninghorizon}$} is submodular, then
\begin{equation}\label{ineq:bound_sub}
\frac{{\metric_{\calV\setminus\calA^\star,\;t+1:t+\planninghorizon}}}{{\optval{\calV}{\attack}}}\geq \max\left(1-\kappa_{{\metric_{\cdot,t+1:t+\planninghorizon}}},\frac{1}{1+\alpha}\right).
\end{equation}

\item{{\emph{Running time:}}} If {$\rho$ upper bounds the {running time}}
for solving the {(attack-free) information acquisition problem} in eq.~\eqref{preq:active_inf_acq},
then \algrobust terminates in $O(|\calV|\rho)$ 
{time}.
\end{itemize}
\end{mytheorem}

{Theorem~\ref{th:per_alg_dec_resil_coord_decent}'s bounds in eqs.~\eqref{ineq:bound_non_sub}-\eqref{ineq:bound_sub} compare \algrobust's selection $u\at{\calV}{1}{t+\planninghorizon}$ against an optimal selection of control inputs that achieves the optimal value $\optval{\calV}{\attack}$ for \prtwo.  Particularly,   eqs.~\eqref{ineq:bound_non_sub}-\eqref{ineq:bound_sub} imply that  for (i) non-decreasing and (ii) non-decreasing and submodular functions $\metric_{\cdot, t+1:t+\planninghorizon}$, \algrobust guarantees a value for \prtwo 
which can be close to the optimal.  
For example,
eq.~\eqref{ineq:bound_sub}'s lower bound $1/(1+\attack)$ is non-zero for any finite number of robots $|\calV|$, and, notably, it equals $1$ in the attack-free case (\algrobust is exact for $\alpha =0$, per Theorem~\ref{th:per_alg_dec_resil_coord_decent}'s assumptions).  
More broadly, 
when  $\kappa_{\metric_{\cdot, t+1:t+\planninghorizon}}<1$ or $c_{\metric_{\cdot, t+1:t+\planninghorizon}}<1$, \algrobust's selection $u\at{\calV}{1}{t+\planninghorizon}$ is close to the optimal, in the sense that Theorem~\ref{th:per_alg_dec_resil_coord_decent}'s bounds are non-zero.
Functions with $\kappa_{\metric_{\cdot, t+1:t+\planninghorizon}}<1$ include the $\log\det$ of {positive-definite} matrices~\cite{sharma2015greedy}; objective functions of this form are the conditional entropy and mutual information when used for batch-state estimation of stochastic processes~\cite{tzoumas2016scheduling}.  Functions with  $c_{\metric_{\cdot, t+1:t+\planninghorizon}}<1$
include the average minimum square error (mean of the trace of a Kalman filter's error covariance across a finite time horizon)~\cite{chamon2017mean}.}  


Theorem~\ref{th:per_alg_dec_resil_coord_decent}'s curvature-dependent bounds in eqs.~\eqref{ineq:bound_non_sub}-\eqref{ineq:bound_sub} also make a first step towards separating
the classes of (i) non-decreasing and (ii) non-decreasing and submodular functions into
functions for which \prtwo
can be approximated well, and functions for which it cannot. Indeed, when either $\kappa_{\metric_{\cdot, t+1:t+\planninghorizon}}$ or $c_{\metric_{\cdot, t+1:t+\planninghorizon}}$ tend to zero, \algrobust becomes exact. 
For example, eq.~\eqref{ineq:bound_non_sub}'s term $1-c_{\metric_{\cdot, t+1:t+\planninghorizon}}$ increases as   $c_{\metric_{\cdot, t+1:t+\planninghorizon}}$ decreases, and its limit is equal to $1$
for $c_{\metric_{\cdot, t+1:t+\planninghorizon}}\rightarrow 0$. {Notably, however, the tightness of Theorem~\ref{th:per_alg_dec_resil_coord_decent}'s bounds is an open problem. For example, although for the attack-free problem in eq.~\eqref{preq:active_inf_acq} a bound $O(1-c_{\metric_{\cdot, t+1:t+\planninghorizon}})$ is known to be optimal (the tightest possible in polynomial time and for a worst-case $\metric_{\cdot, t+1:t+\planninghorizon}$)~\cite[Theorem~8.6]{sviridenko2017optimal},
the optimality of eq.~\eqref{ineq:bound_non_sub} is an open problem.}

Overall, Theorem~\ref{th:per_alg_dec_resil_coord_decent} quantifies \algrobust's approximation performance when the robots in~$\calV$ solve optimally the (attack-free) information acquisition problems in \algrobust's line~\ref{line1:opt_1}, line~\ref{line1:step_3}, and line~\ref{line1:step_4}. Among those, however, the problems in line~\ref{line1:step_3} and line~\ref{line1:step_4} are computationally challenging, being multi-robot coordination problems; only approximation algorithms are known for their solution.  Such an approximation algorithm is the recently proposed \emph{coordinate descent}~\cite[Section~IV]{atanasov2015decentralized}.  Coordinate descent has the advantages of  having a provably near-optimal approximation performance.  Therefore, we next quantify \algrobust's performance when the robots in~$\calV$  solve the problems in \algrobust's line~\ref{line1:step_3}, and line~\ref{line1:step_4} using {coordinate descent}.\footnote{{We refer to \ref{app:description_coordinate} for a description of {coordinate descent}.}}

\begin{myproposition}[Approximation Performance of \algrobust via {Coordinate Descent}]\label{prop:alg_per_with_coordinate_descent}
Consider an instance of \prtwo.  Assume the robots in~$\calV$ 
solve the {(attack-free) information acquisition problem in eq.~\eqref{preq:active_inf_acq} 
suboptimally in the case of multiple robots \emph{($|\calV|\geq 2$)} via {coordinate descent}~\cite[Section~IV]{atanasov2015decentralized}, and opti- mally in the case of a single robot \emph{($|\calV|=1$)}.} Then:

\begin{itemize}
\item{\emph{Approximation performance:}} 
\algrobust returns control inputs {$u\at{\calV}{1}{t+\planninghorizon}$} such that (i) if {$\metric_{\cdot, t+1:t+\planninghorizon}:2^\calV\mapsto \mathbb{R}$} is non-decreasing, and, without loss of generality, normalized and non-negative, then
    \begin{equation}\label{ineq:bound_non_sub_cd}
	 \frac{
	 {
	 \metric_{\calV\setminus\calA^\star,\;t+1:t+\planninghorizon}
	 }
	 }{{\optval{\calV}{\attack}}}\geq \frac{1}{2}(1-c_{{\metric_{\cdot, t+1:t+\planninghorizon}}})^3;
	\end{equation}

(ii) If {$\metric_{\cdot, t+1:t+\planninghorizon}$} is also submodular, then
\begin{equation}\label{ineq:bound_sub_cd}
\hspace{-4mm}\frac{{\metric_{\calV\setminus\calA^\star,\;t+1:t+\planninghorizon}}}{{\optval{\calV}{\attack}}}\geq\frac{1}{2} \max\left(1-\kappa_{{\metric_{\cdot,t+1:t+\planninghorizon}}},\frac{1}{1+\alpha}\right).
\end{equation}

{\item{{\emph{Running time:}}} If $\rho_{\scenario{CD}}$ upper bounds the {running time}
for solving the information acquisition problem in eq.~\eqref{preq:active_inf_acq} via coordinate descent,
then \algrobust terminates in $O(\rho_{\scenario{CD}})$ 
{time}.
}
\end{itemize}
\end{myproposition}

{Proposition~\ref{prop:alg_per_with_coordinate_descent}'s suboptimality bounds are discounted versions of Theorem~\ref{th:per_alg_dec_resil_coord_decent}'s bounds: (i) eq.~\eqref{ineq:bound_non_sub_cd} is the discounted eq.~\eqref{ineq:bound_non_sub} by the factor $(1-c_{\metric_{\cdot, t+1:t+\planninghorizon}})/2$; and (ii)  eq.~\eqref{ineq:bound_sub_cd} is the discounted eq.~\eqref{ineq:bound_sub} by the factor $1/2$.  The source of the discounting factors is the requirement in Proposition~\ref{prop:alg_per_with_coordinate_descent} that the robots in $\calV$ can solve only suboptimally (via coordinate descent) the information acquisition problem in eq.~\eqref{preq:active_inf_acq} (and, in effect, the problems in \algrobust's line~\ref{line1:step_3} and line~\ref{line1:step_4}).  In more detail, in Lemma~\ref{lem:generalized_cd}, located in \ref{app:description_coordinate}, we prove that  (i) for non-decreasing objective functions, coordinate descent guarantees the suboptimality bound $(1-c_{\metric_{\cdot, t+1:t+\planninghorizon}})/2$ for eq.~\eqref{preq:active_inf_acq} (which is the discounting factor to eq.~\eqref{ineq:bound_non_sub}, resulting in eq.~\eqref{ineq:bound_non_sub_cd}), while (ii) for non-decreasing and submodular functions, coordinate descent is known to guarantee the suboptimality bound $1/2$ for eq.~\eqref{preq:active_inf_acq} (which is the discounting factor to eq.~\eqref{ineq:bound_sub}, resulting in eq.~\eqref{ineq:bound_sub_cd})~\cite{atanasov2015decentralized}.
}

{Proposition~\ref{prop:alg_per_with_coordinate_descent} also implies that if the robots in $\calV$ use coordinate descent to solve the (attack-free) information acquisition problems in \algrobust's line~\ref{line1:step_3} and line~\ref{line1:step_4}, then \algrobust has the same of order of running time as coordinate descent.  The proof of  Proposition~\ref{prop:alg_per_with_coordinate_descent} is found in \ref{app:proof_proposition}.
}

\section{Applications and Experiments}\label{sec:simulations_experiments}
We present \myalg's performance  in applications.
We present three applications of \textit{Resilient Active Information Acquisition with Teams of Robots}: (i) \emph{Multi-Target Tracking} {(Section~\ref{subsec:multi-target})}, (ii) \emph{Occupancy Grid Mapping} {(Section~\ref{subsec:occupancy})}, and (iii) \emph{Persistent surveillance} {(Section~\ref{subsec:persistent})}. 
{We confirm \myalg effectiveness, even as we vary key parameters in \prone: \begin{enumerate}
    \item[(i)] the \emph{number of attacks $\alpha$}, among the permissible values  $\{0,1,\ldots, |\calV|\}$, to test \myalg's performance to both small and high attack numbers {(Section~\ref{subsec:multi-target})};
    \item[(ii)] the \emph{attack model}, beyond the worst-case model prescribed by \prone's problem formulation, to test \myalg's sensitivity against \textit{non} worst-case failures; particularly, random failures {(Section~\ref{subsec:occupancy})}.\footnote{For random failures, one would expect {\sf RAIN}'s performance to be the same, or improve, since {\sf RAIN} is designed to withstand the worst-case.}
    \item[(iii)] the \emph{replanning rate $\replanrate$}, among the permissible values $\{1,2,\ldots, {\planninghorizon}\}$, to test \myalg's performance even when the replanning rate is high {(Section~\ref{subsec:persistent})}.\footnote{{When the replanning rate tends, ideally, to infinity, in that the robots can instantaneously observe all attacks and replan (which, however, is in practice impossible due to algorithmic, computational, and communication delays), then it is expected that {\sf \footnotesize RAIN}'s advantage over a non-resilient algorithm with the same replanning rate, such as coordinate descent, would diminish.\label{foot:replan}}}
\end{enumerate}
}

{
\textbf{Common Experimental Setup across Applications:}
}
{\paragraph{Approximation Algorithm for (Attack-Free) Information Acquisition problem in \emph{eq.~\eqref{preq:active_inf_acq}} (and in effect for the problems in \algrobust's line~\ref{line1:opt_1}, line~\ref{line1:step_3}, and line~\ref{line1:step_4})} In the multi-robot case (pertained to \algrobust's line~\ref{line1:step_3}, and line~\ref{line1:step_4}), the algorithm used for approximating a solution to eq.~\eqref{preq:active_inf_acq} is the \emph{coordinate descent}~\cite{atanasov2015decentralized} (also, cf.~\ref{app:description_coordinate}).  Evidently, coordinate descent does not account for the possibility of attacks, and for this reason, we also use it as a baseline to compare \myalg with. In the single robot case (pertained to \algrobust's line~\ref{line1:opt_1}),  eq.~\eqref{preq:active_inf_acq} reduces to a single-robot motion planning problem, and for its solution we use reduced value iteration {(\scenario{ARVI} algorithm \cite{schlotfeldt2018anytime})}, except for the application of \textit{Occupancy Grid Mapping} (Section~\ref{subsec:occupancy}) where we use forward value iteration {\cite{atanasov2014information}}}.

\paragraph{{Worst-Case Attack Approximation}}
{Computing the worst-case attack requires brute-force, since the minimization step in \prone is NP-hard~\cite{orlin2016robust}.  The consequence is that solving for the worst-case attack requires solving an exponential number of instances of the information acquisition problem in eq.~\eqref{preq:active_inf_acq}, prohibiting real-time navigation performance by the robots, {even for small  teams of robots ($|\calV|\geq 5$)}. In particular, the running time required to solve eq.~\eqref{preq:active_inf_acq}, even via coordinate descent, can be exponential in the number of robots and task length horizon, namely,  $\mathcal{O}(|\mathcal{U}|^{|\calV| \missionlength})$~\cite{atanasov2015decentralized}  ($\mathcal{U}$ denotes the set of admissible control inputs to each of the robots in $\calV$, assumed the same across all robots).  Hence, we approximate the worst-case attacks by solving the minimization step in \prone via {a greedy algorithm~\cite{nemhauser1978analysis}}.}

\paragraph{Computational Platform} {Experiments are implemented in C++, and run on an Intel Core i7 CPU laptop.}

\subsection{Resilient Multi-Target Tracking}\label{subsec:multi-target}

\begin{figure}
\centering
\includegraphics[width=0.75\linewidth]{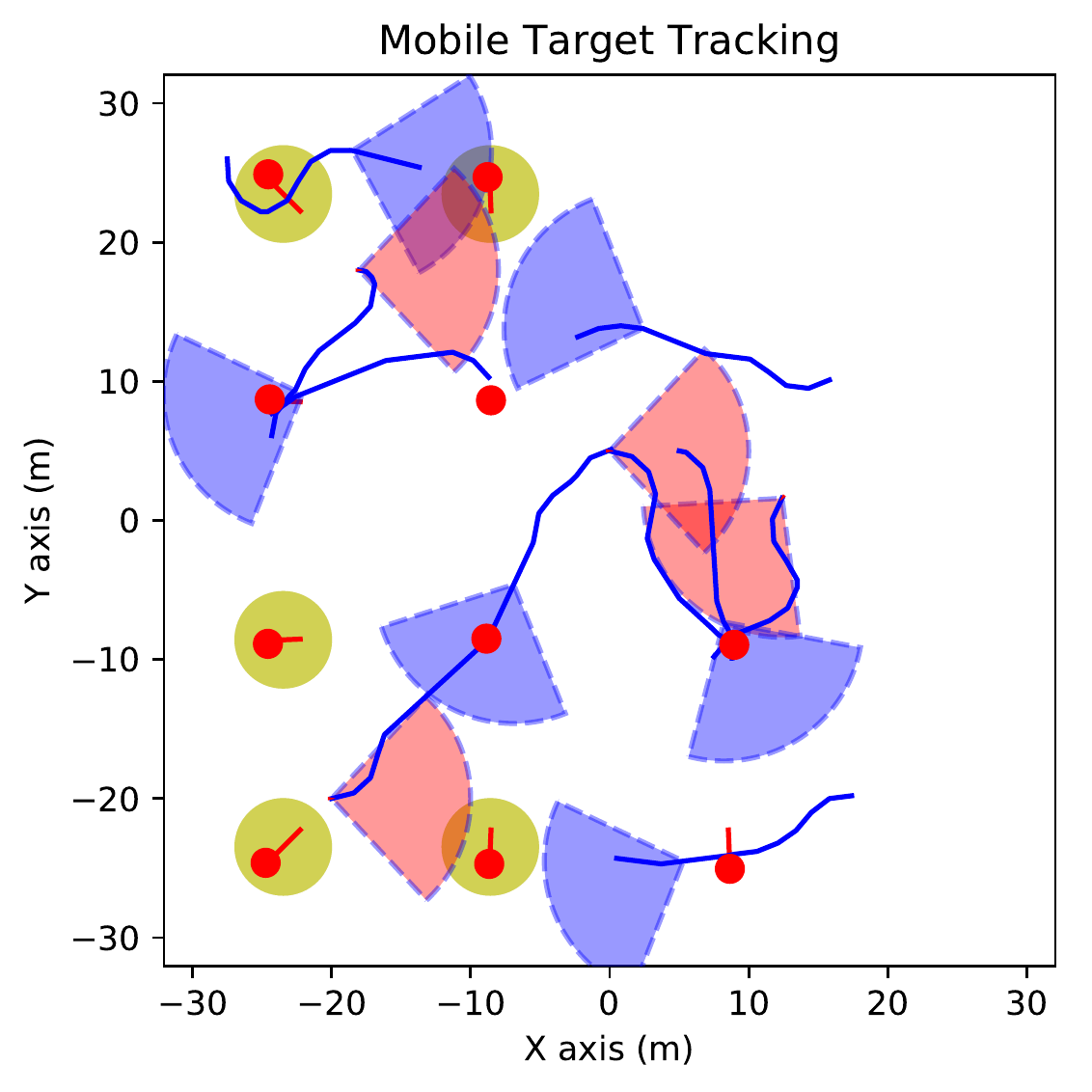}
   \vspace{-2mm}
\caption{\textbf{Resilient Multi-Target Tracking Scenario}.
10 robots are depicted tracking 10 targets, while 4 of the robots are be attacked (causing their sensing capabilities to be, at least temporarily, disabled).  The robots are depicted with their conic-shaped field-of-view, colored light blue for non-attacked robots and light red for attacked robots. 
The targets are depicted with red disks.
Planned robot trajectories are shown as solid blue lines.  Predicted target trajectories are shown as solid red lines.
Each light-green ellipse represents the covariance of the target's location estimate.}
  \label{fig:sim_ex}
\end{figure}

In \emph{Resilient Multi-Target Tracking}, a team of mobile robots is tasked to track the locations of multiple moving targets, even in the presence of a number of attacks against the robots. {For the purpose of assessing \myalg's effectiveness against various number of attacks, we will vary the number of attacks across scenarios, where we will also possibly vary the number of robots and targets.  In more detail, the experimental setup and simulated scenarios are described below.} 

\myParagraph{Experimental Setup}  {We specify the used (a) \emph{robot dynamics}, (b) \emph{target process}, (c) \emph{sensor model}, and (d) \emph{information acquisition objective function}:} 
\setcounter{paragraph}{0}

\paragraph{Robot Dynamics} Each robot $i$ has unicycle dynamics in \SEtwo, discretized with a sampling period $\tau$, such that 
\begin{align}
\label{eq:dynamics}
\scaleMathLine[.88]{
\begin{pmatrix}
x_{i,t+1}^{(1)} \\ x_{i,t+1}^{(2)} \\ \theta_{i,t+1}
\end{pmatrix} = 
\begin{pmatrix}
x_{i,t}^{(1)} \\x_{i,t}^{(2)} \\ \theta_{i,t} 
\end{pmatrix} + 
\begin{pmatrix}
\nu_i \sinc(\frac{\omega_i \tau}{2}) \cos (\theta_{i,t} + \frac{\omega_i \tau}{2})\\
\nu_i \sinc(\frac{\omega_i \tau}{2}) \sin (\theta_{i,t} + \frac{\omega_i \tau}{2})\\
\tau \omega_i
\end{pmatrix},}
\end{align}
where $(\nu_i,\omega_i)$ is the control input (linear and angular velocity). 

\paragraph{Target Dynamics} The targets move according to double integrator dynamics, which are assumed corrupted with additive Gaussian noise.
Specifically, if $M$ denotes the number of targets, then $y_t = [y_{t,1}^\top, \ldots, y_{t,M}^\top]^\top$,
where $y_{t,m}$ is the planar coordinates and velocities of target $m$, and
\begin{align*}
\scaleMathLine[1]{
y_{t+1,m} = \begin{bmatrix} I_2 &\tau I_2 \\
					0 & I_2  \end{bmatrix}y_{t,m} + w_t, \hspace{3mm} w_t \sim \setN\begin{pmatrix}0,q \begin{bmatrix} \tau^3/3 I_2 & \tau^2/2 I_2 \\ \tau^2/2 I_2 & \tau I_2 \end{bmatrix}\end{pmatrix}.}                    
\end{align*}
with $q$ being a noise diffusion parameter.

\paragraph{Sensor Model} The robots' sensor model consists of a range and bearing for each target $m =1,\ldots, M$:
\begin{equation*}
\begin{aligned}
&\scaleMathLine[0.7]{z_{t,m} = h(x_t,y_{t,m}) + v_t, \hspace{3mm} v_t \sim \setN(0, V(x_t,y_{t,m}));}\\
&\scaleMathLine[0.88]{h(x,y_m) = \begin{bmatrix}r(x,y_m)\\\alpha(x,y_m)\end{bmatrix} \triangleq \begin{bmatrix} \sqrt{(y^1 - x^1)^2 + (y^2 -x^2)^2} \\ \tan^{-1} ((y^2 - x^2)(y^1 - x^1)) - \theta \end{bmatrix}.}
\end{aligned} 
\end{equation*}
Since the sensor model is non-linear, we linearize it around the currently predicted target trajectory. Particularly, given
\begin{align*}
\scaleMathLine[1]{\nabla_y h(x,y_m) = \frac{1}{r(x,y_m)} 
\begin{bmatrix} (y^1 - x^1) & (y^2 - x^2) & 0_{1x2} \\
-\sin( \theta + \alpha(x,y_m)) & \cos (\theta + \alpha(x,y_m)) & 0_{1x2} \\ \end{bmatrix},}
\end{align*}
the observation model for the joint target state can be expressed as a block diagonal matrix containing the linearized observation models for each target along the diagonal:
\begin{align*}
H \triangleq \diag{\nabla_{y_1}h(x,y_1), \ldots, \nabla_{y_M}h(x,y_M)}.
\end{align*}
The sensor noise covariance grows linearly in range and in bearing, up to $\sigma_r^2$, and $\sigma_b^2$, where $\sigma_r$ and $\sigma_b$ are the standard deviation of the range and the bearing noise, respectively. The model here also includes a limited range and field of view, denoted by the parameters $r_{sense}$ and $\psi$, respectively.

\begin{figure}[t]
\captionsetup[subfloat]{labelformat=empty}
\vspace{-4mm}
\hspace{-4mm}
      \subfloat[][]{ \includegraphics[width=0.5\columnwidth]{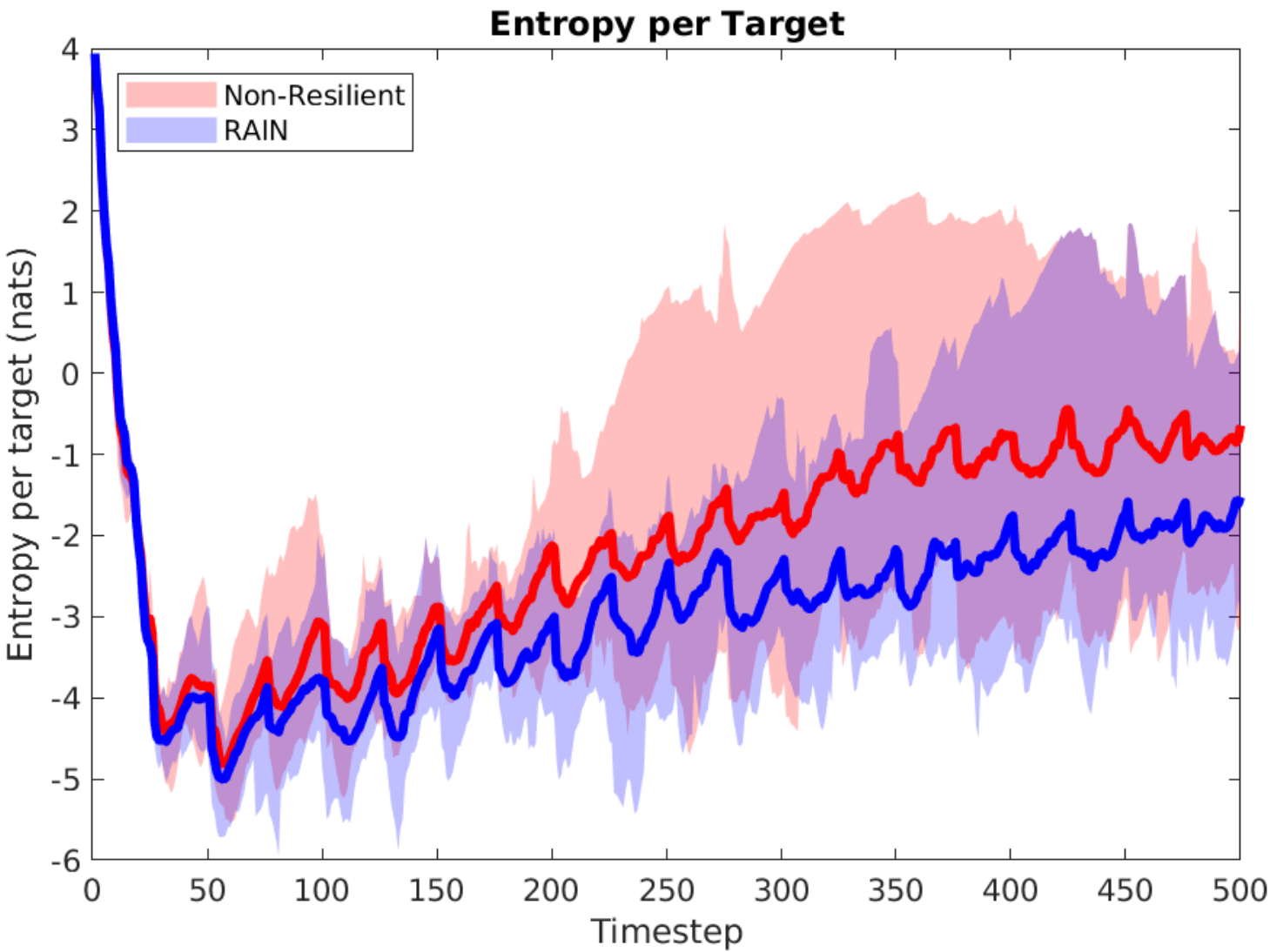}}
   \subfloat[][]{ \includegraphics[width=0.5\columnwidth]{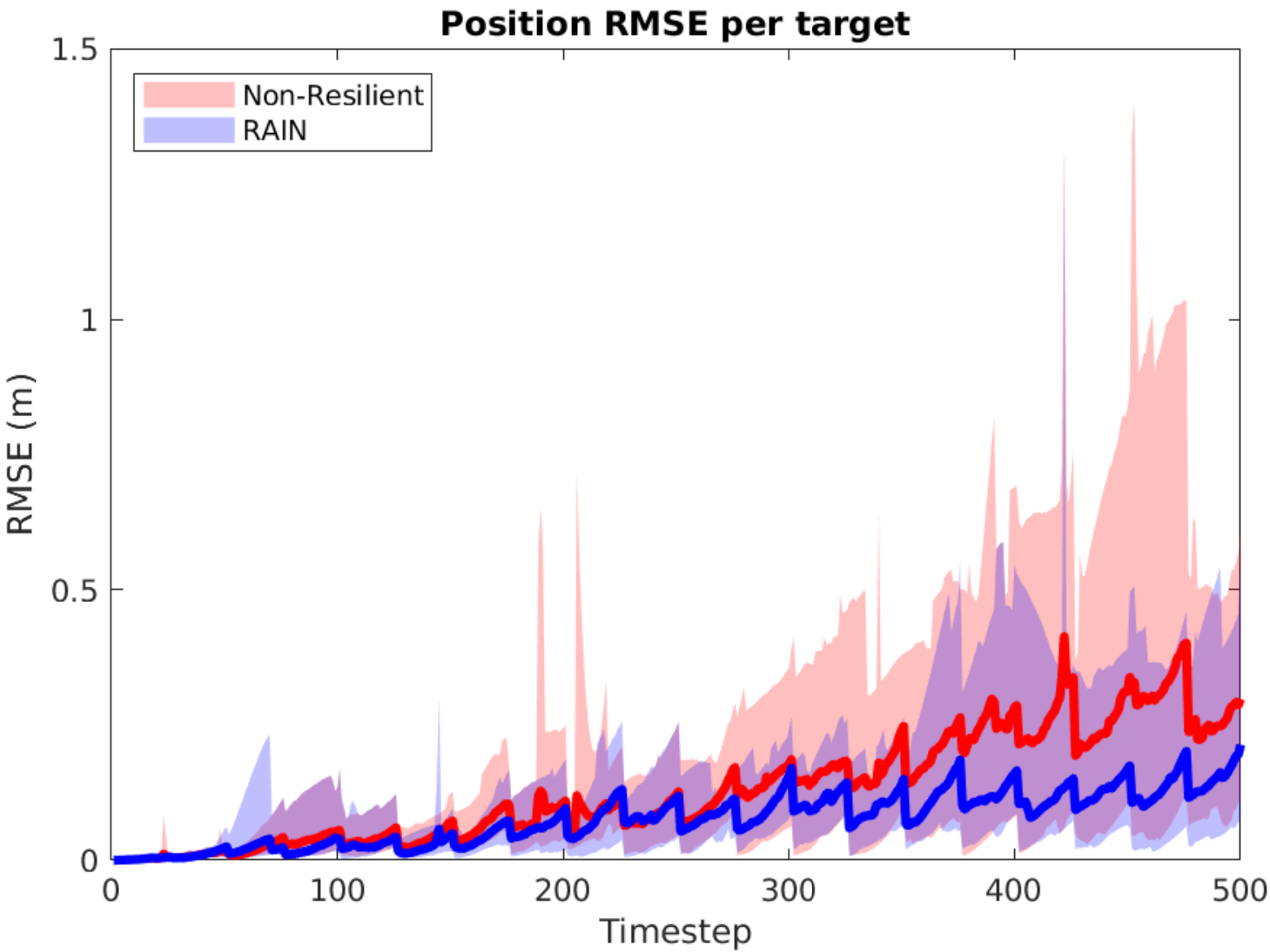}} 
      \vspace{-3ex}\\
   \vspace{-3ex}
   \hspace{-4mm}
      \subfloat[][]{ \includegraphics[width=0.5\columnwidth]{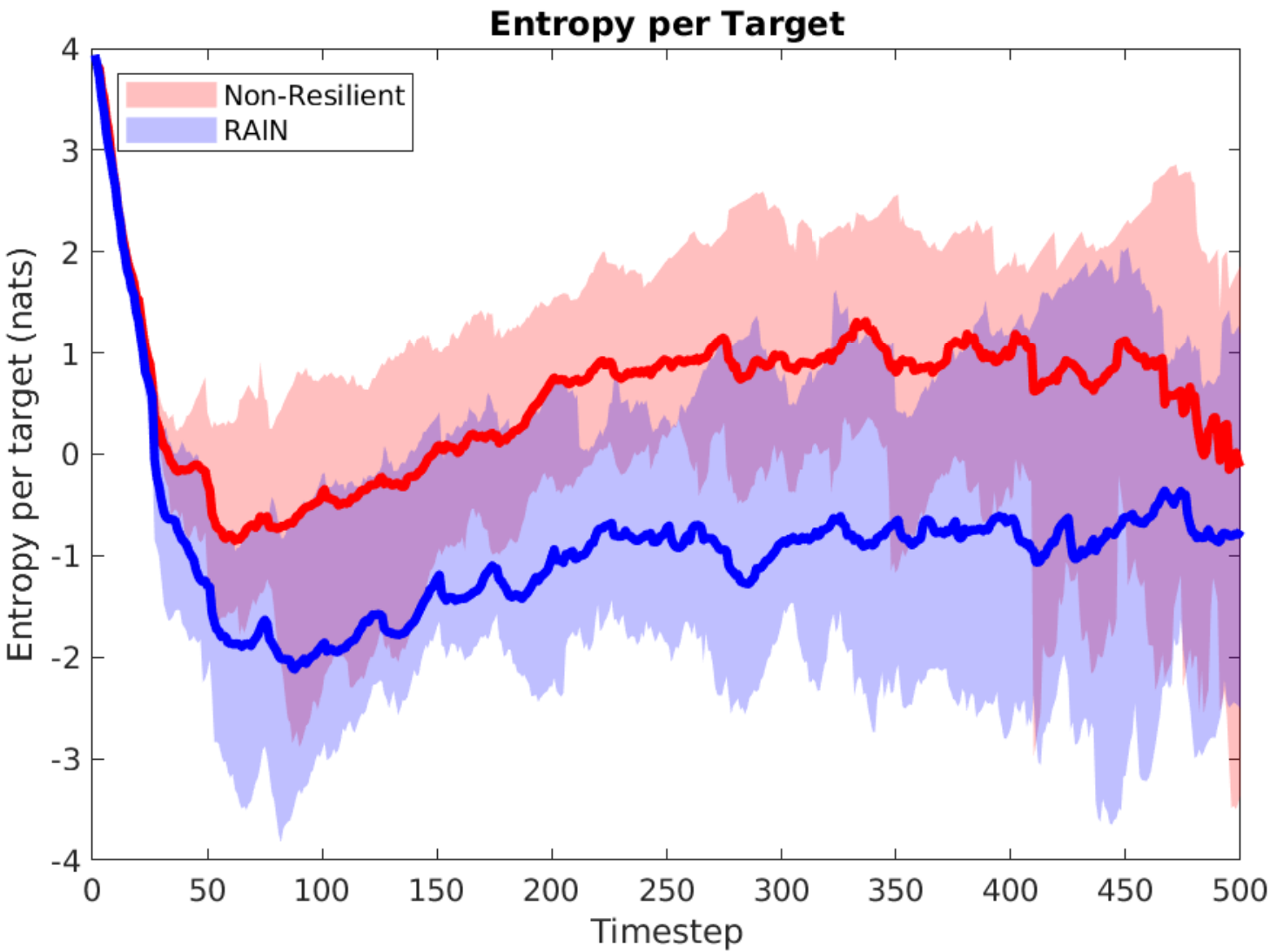}} 
         \subfloat[][]{ \includegraphics[width=0.5\columnwidth]{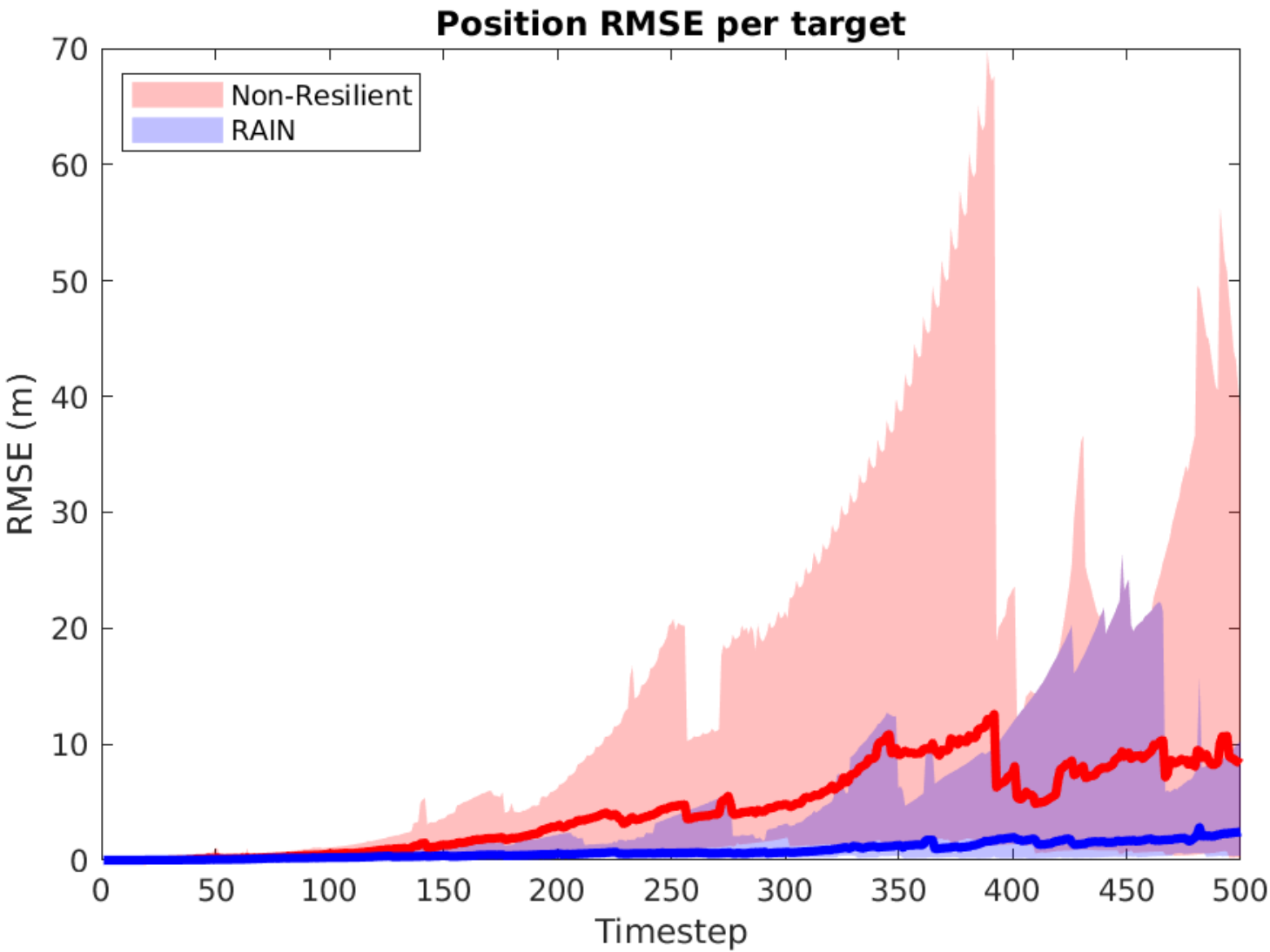}}
   \vspace{1mm}
   \caption{
   \textbf{Resilient Multi-Target Tracking Results.} 
  Performance comparison of {\sf \footnotesize RAIN} with coordinate descent (noted as {NonResilient in the plots}), for two configurations and two performance metrics:  top row considers 10 robots, 10 targets, and 2 attacks; bottom row considers the same number of robots and targets but 6 attacks.  The left column depicts mean entropy per target, averaged over the robots; the right column depicts the position Root Mean Square Error ({RMSE}) per target, also averaged over the robots.}
      \label{fig:sim_plots}
   \end{figure}

\paragraph{Information Acquisition Objective Function} For the information objective function we use the time averaged log determinant of the covariance matrix, which is equivalent to conditional entropy for Gaussian variables\cite{atanasov2014information}. This objective function is non-decreasing, yet not necessarily submodular~\cite{jawaid2015submodularity}. 
Overall, we solve an instance of \prone per the aforementioned setup and the objective function:
\begin{equation*}
J_{\calV, \;1:\missionlength} \triangleq \frac{1}{{\missionlength}}\sum_{t=1}^{{\missionlength}} \log \det (\Sigma_{\calV, \;t}), \nonumber
\end{equation*}
where $\Sigma_{\calV,\;t}$ is the Kalman filtering error covariance at $t$, given the measurements collected up to $t$ by the robots in $\calV$~\cite{atanasov2014information}.\footnote{The multi-target tracking scenarios are dependent on a prior distribution of the target's initial conditions $y_{0}$ and $\Sigma_{0|0}$, assumed here known. Yet, if a prior distribution is unknown, then an exploration strategy can be incorporated to find the targets by placing exploration landmarks at the map frontiers \cite{atanasov2015decentralized}.}  


\myParagraph{Simulated Scenarios} We consider multiple scenarios of the experimental setup introduced above: across scenarios, we
vary the number of robots, $n$, the number of targets $M$, and the number of attacks, $\alpha$; cf.~first column of Table~\ref{fig:sim_table}. Additionally, the robots and targets are restricted to move inside a $64\times64 m^2$ environment (Fig.~\ref{fig:sim_ex}).  The admissible control input values to each robot are the $\mathcal{U} = \{1, 3\}m/s \times \{0, \pm 1, \pm 3\}rad/s$.
At the beginning of each scenario, we fix the initial positions of both the robots and targets, and the robots are given a prior distribution of the targets before starting the simulation. The targets start with a zero velocity, and in the event that a target leaves the environment its velocity is reflected to remain in bounds. Finally, across all simulations:  {$\planninghorizon=\replanrate=25$} steps, $\tau= 0.5 s$, $r_{sense}=10 m$, $\psi=94^{\circ}$, $\sigma_r=.15 m$, $\sigma_b = 5^{\circ}$, and $q=.001$. We use $\missionlength=500$.
\myParagraph{Compared Techniques} We compare \myalg with \textit{coordinate descent}. We consider two performance measures: the average entropy, and average Root Mean Square Error (RMSE) per target, averaged over the robots in the team.

\begin{table}[t]
\captionsetup{font=small}
\centering
\label{my-label}
\begin{tabular}{l|cc|cc}
    \cline{2-5}                                                   \multirow{2}{*}{}&\multicolumn{2}{c|}{Mean  RMSE} & \multicolumn{2}{c|}{Peak RMSE}
                                                        \\ \cline{2-5}
                                 & \scenario{NonRes} & \cellcolor[HTML]{C0C0C0}\myalg & \scenario{NonRes} & \multicolumn{1}{l|}{\cellcolor[HTML]{C0C0C0}\myalg} \\ \cline{1-5}
\rowcolor[HTML]{EFEFEF} 
\multicolumn{1}{|l|}{\cellcolor[HTML]{EFEFEF}$n=5$, $M=10$} & \multicolumn{2}{l|}{\cellcolor[HTML]{EFEFEF}} & \multicolumn{2}{l|}{\cellcolor[HTML]{EFEFEF}}                     \\
\multicolumn{1}{|l|}{$\alpha=1$}                        & 0.28     & \cellcolor[HTML]{C0C0C0}0.19      & 9.62     & \multicolumn{1}{l|}{\cellcolor[HTML]{C0C0C0}2.09}      \\
\multicolumn{1}{|l|}{$\alpha=2$}                        & 1.47     & \cellcolor[HTML]{C0C0C0}0.68      & 26.07    & \multicolumn{1}{l|}{\cellcolor[HTML]{C0C0C0}15.71}     \\
\multicolumn{1}{|l|}{$\alpha=4$}                        & 10.67    & \cellcolor[HTML]{C0C0C0}4.9       & 225.47   & \multicolumn{1}{l|}{\cellcolor[HTML]{C0C0C0}103.82}    \\
\rowcolor[HTML]{EFEFEF} 
\multicolumn{1}{|l|}{\cellcolor[HTML]{EFEFEF}$n=10$, $M=5$}  & \multicolumn{2}{l|}{\cellcolor[HTML]{EFEFEF}} & \multicolumn{2}{l|}{\cellcolor[HTML]{EFEFEF}}                     \\
\multicolumn{1}{|l|}{$\alpha=2$}                        & 0.35     & \cellcolor[HTML]{C0C0C0}0.14      & 57.65    & \multicolumn{1}{l|}{\cellcolor[HTML]{C0C0C0}1.87}      \\
\multicolumn{1}{|l|}{$\alpha=4$}                        & 0.39     & \cellcolor[HTML]{C0C0C0}0.28      & 6.66     & \multicolumn{1}{l|}{\cellcolor[HTML]{C0C0C0}3.17}      \\
\multicolumn{1}{|l|}{$\alpha=6$}                        & 2.07     & \cellcolor[HTML]{C0C0C0}0.65      & 93.27    & \multicolumn{1}{l|}{\cellcolor[HTML]{C0C0C0}15.63}     \\
\rowcolor[HTML]{EFEFEF} 
\multicolumn{1}{|l|}{\cellcolor[HTML]{EFEFEF}$n=10$, $M=10$} & \multicolumn{2}{l|}{\cellcolor[HTML]{EFEFEF}} & \multicolumn{2}{l|}{\cellcolor[HTML]{EFEFEF}}                     \\
\multicolumn{1}{|l|}{$\alpha=2$}                        & 0.13     & \cellcolor[HTML]{C0C0C0}0.08      & 1.4      & \multicolumn{1}{l|}{\cellcolor[HTML]{C0C0C0}1.32}      \\
\multicolumn{1}{|l|}{$\alpha=4$}                        & 0.24     & \cellcolor[HTML]{C0C0C0}0.23      & 4.19     & \multicolumn{1}{l|}{\cellcolor[HTML]{C0C0C0}2.66}      \\
\multicolumn{1}{|l|}{$\alpha=6$}                        & 4.39     & \cellcolor[HTML]{C0C0C0}1.2       & 69.77    & \multicolumn{1}{l|}{\cellcolor[HTML]{C0C0C0}26.4}      \\ \hline
\end{tabular}
\vspace{1mm}
\caption{\textbf{Resilient Multi-Target Tracking Results.} Performance comparison of {\sf \footnotesize RAIN} with coordinate descent (noted as {{\sf \footnotesize NonRes} in the table}), for a variety of configurations, where $n$ denotes the number of robots ($n=|\calV|$), $M$ denotes the number of targets, and $\alpha$ denotes the number of failures. Two performance metrics are used: mean Root Mean Square Error (RMSE), and peak RMSE, both per target, and averaged over the robots in the team.}
 \label{fig:sim_table}
\end{table}

\begin{figure*}[t]
   \captionsetup[subfloat]{labelformat=empty}
    \centering
   \hspace{-6mm}\subfloat[][]{ \includegraphics[width=.6\columnwidth]{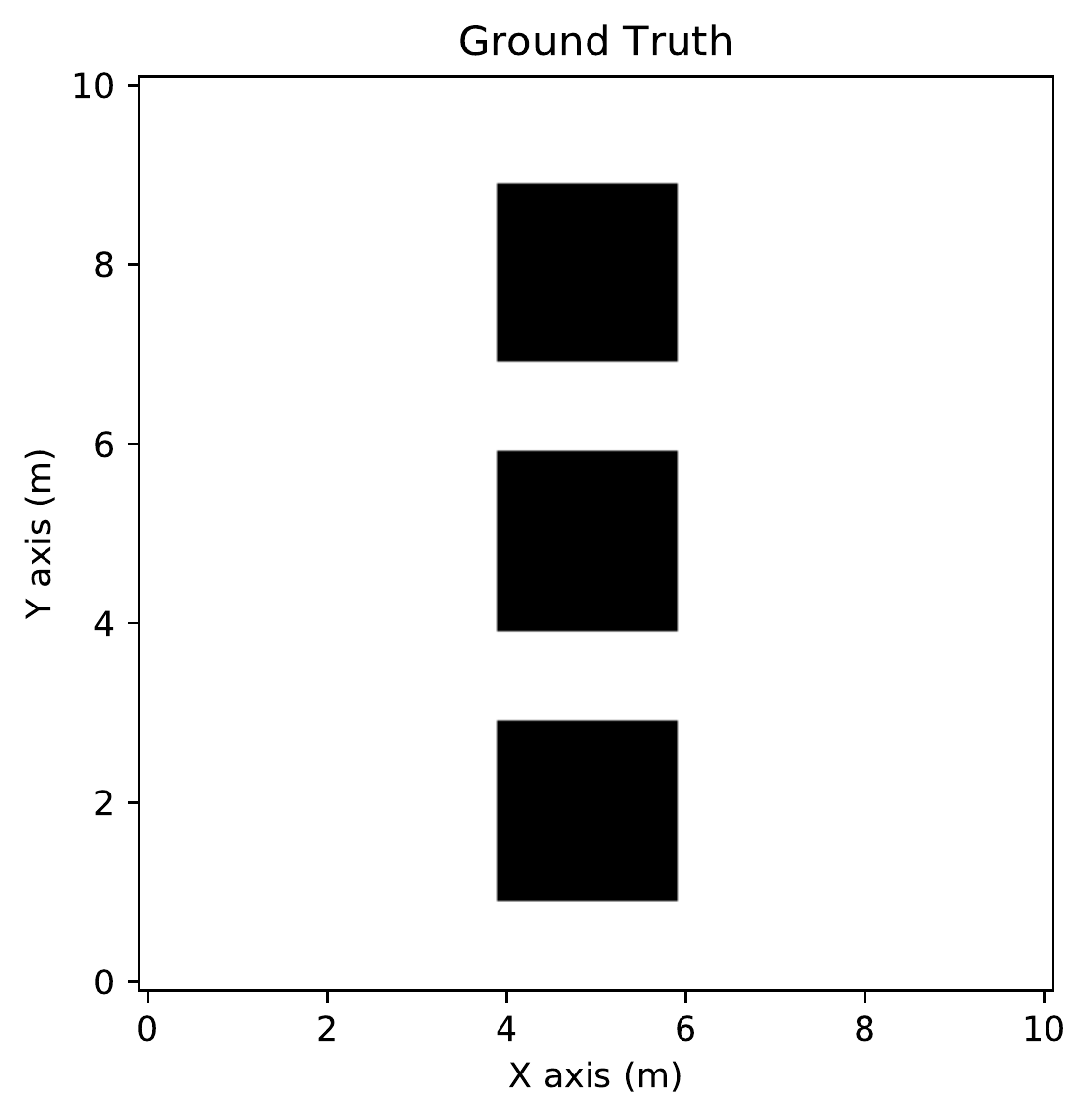}}~~
      \subfloat[][]{ \includegraphics[width=.6\columnwidth]{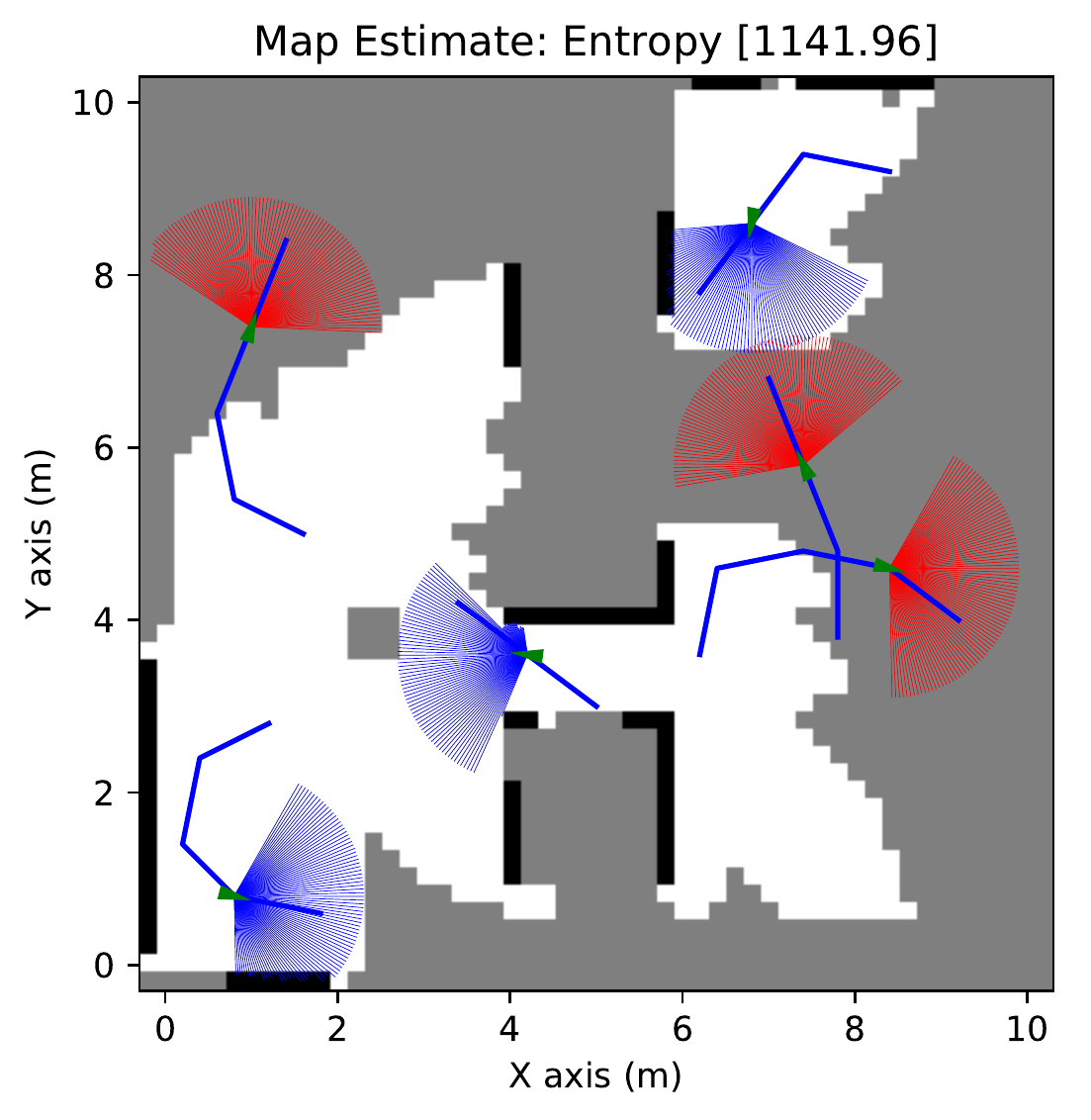}} ~~
      \subfloat[][]{ \includegraphics[width=.6\columnwidth]{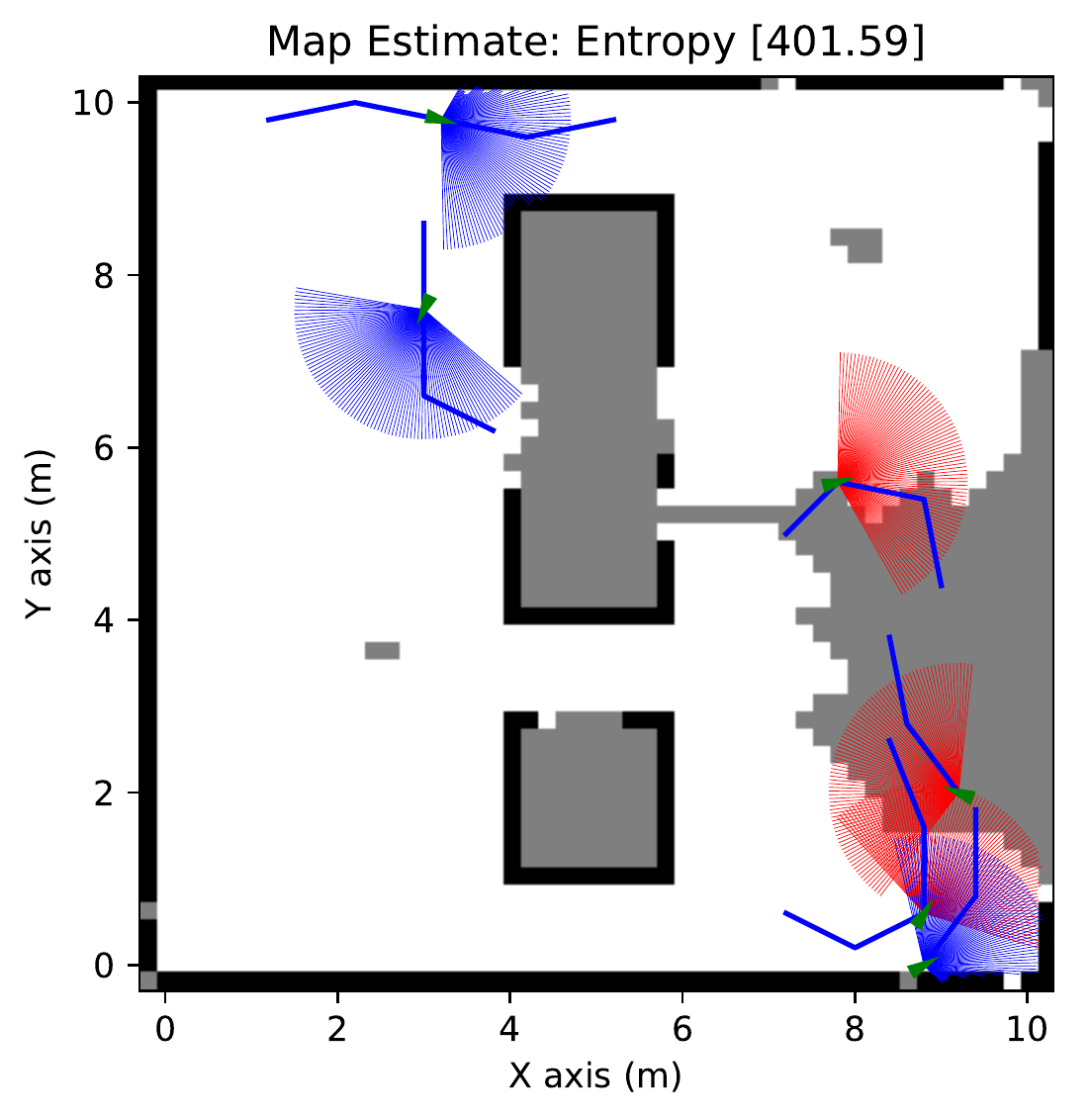}} \\
      \vspace{-4ex}
         \hspace{-6mm}
   \subfloat[][]{ \includegraphics[width=.6\columnwidth]{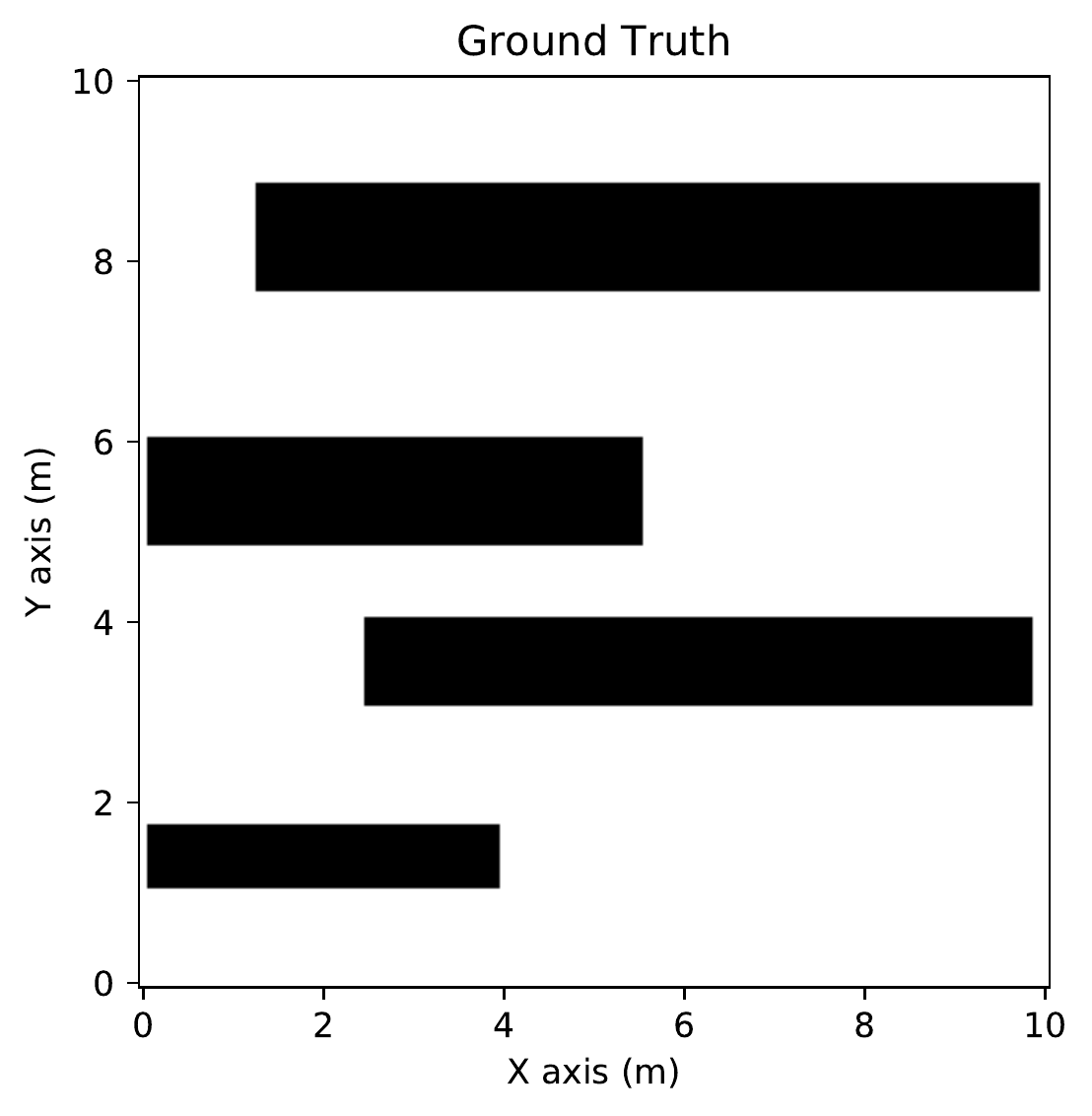}}~~ 
   \subfloat[][]{ \includegraphics[width=.6\columnwidth]{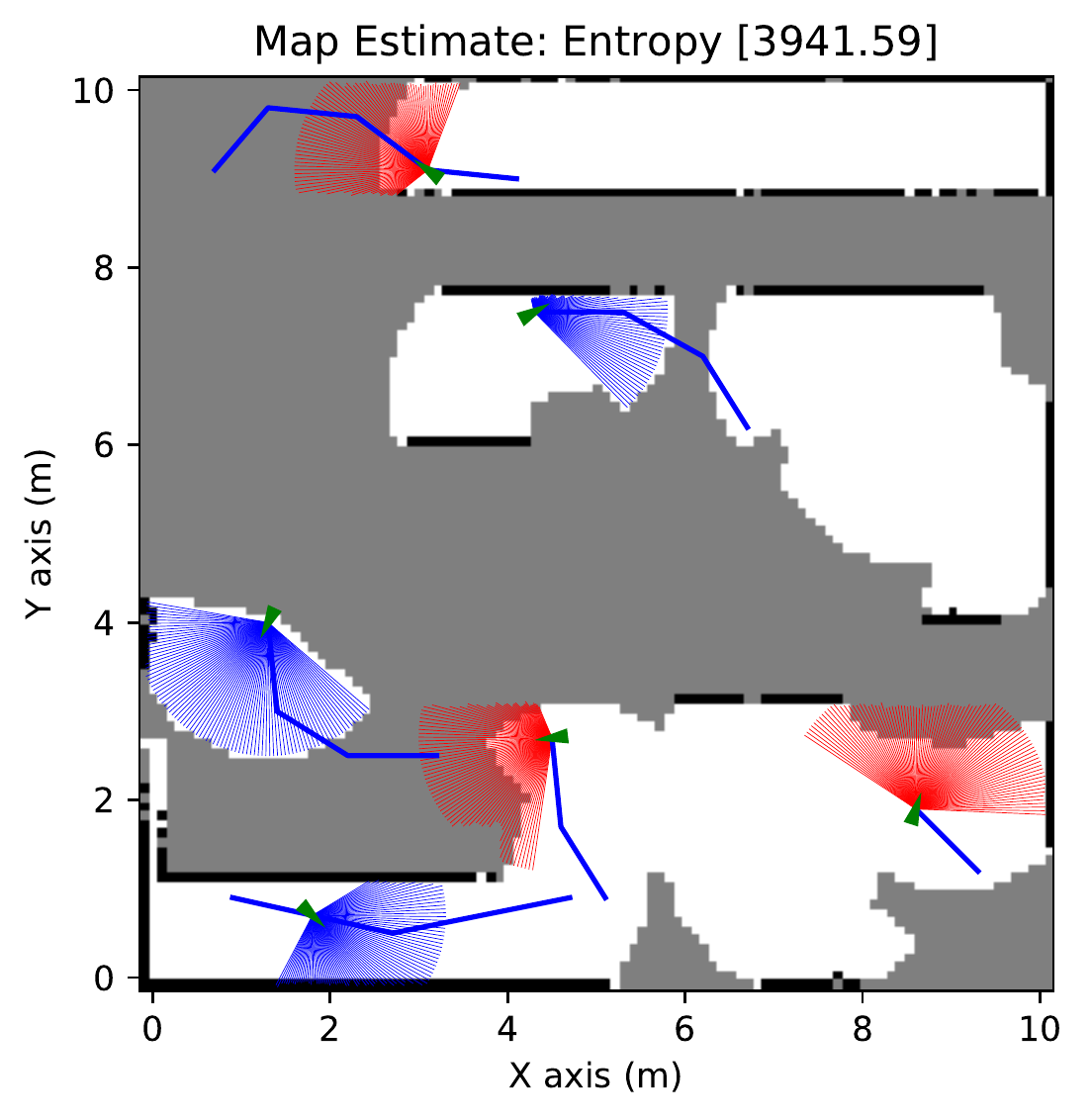}}~~
   \subfloat[][]{ \includegraphics[width=.6\columnwidth]{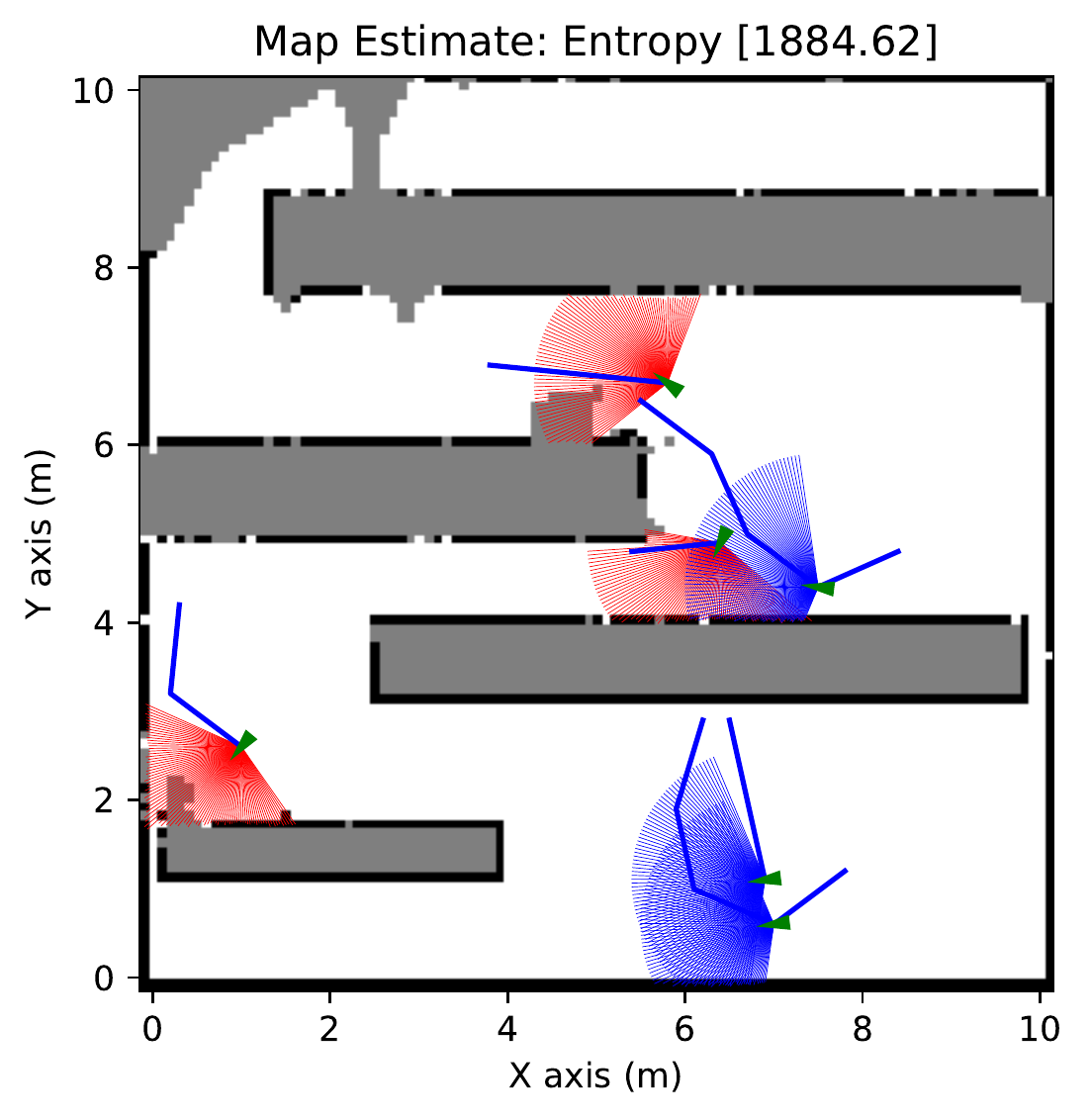}}
   \vspace{-6mm}
   \caption{\textbf{Resilient Occupancy Grid Mapping Scenarios}. Two scenarios are considered:
   a square obstacle map (top row), and a corridor map (bottom row), where free space is colored white, occupied space is colored black, and unexplored/unknown space is colored gray. The non-attacked robots are shown with their field-of-view colored blue, whereas the attacked robots are shown with field-of-view colored red. The left-most column shows the considered ground truth maps; the middle column shows the map estimate halfway through the task horizon; the right-most column shows the map estimate near completion.}
      \label{fig:sim_plots_occ}
   \end{figure*}
   
   \myParagraph{Results} The results, averaged across {10 Monte-Carlo runs}, are depicted in Fig.~\ref{fig:sim_plots} and Table~\ref{fig:sim_table}. In Fig. \ref{fig:sim_plots}, \myalg's performance is observed to be superior both in terms of the average entropy and the RMSE.  Particularly, as the number of attacks grows (cf.~second rows of plots in Fig. \ref{fig:sim_plots}), \myalg's benefits are accentuated in comparison to the non-resilient coordinate descent.
Similarly, Table~\ref{fig:sim_table} demonstrates that \myalg achieves a lower mean RMSE than coordinate descent, and, crucially, is highly effective in reducing the peak estimation error; in particular, \myalg achieves a performance that is 2 to 30 times better in comparison to the performance achieved by the non-resilient algorithm. 
We also observe that the impact of Algorithm~\ref{alg:dec_resil_coord_decent} is most prominent when the number of attacks is large relative to the size of the robot team.
   
\begin{figure*}[t]
   \captionsetup[subfloat]{labelformat=empty}
    \centering
 \subfloat[][]{     \hspace{-5mm}\includegraphics[width=280px,height=165px]{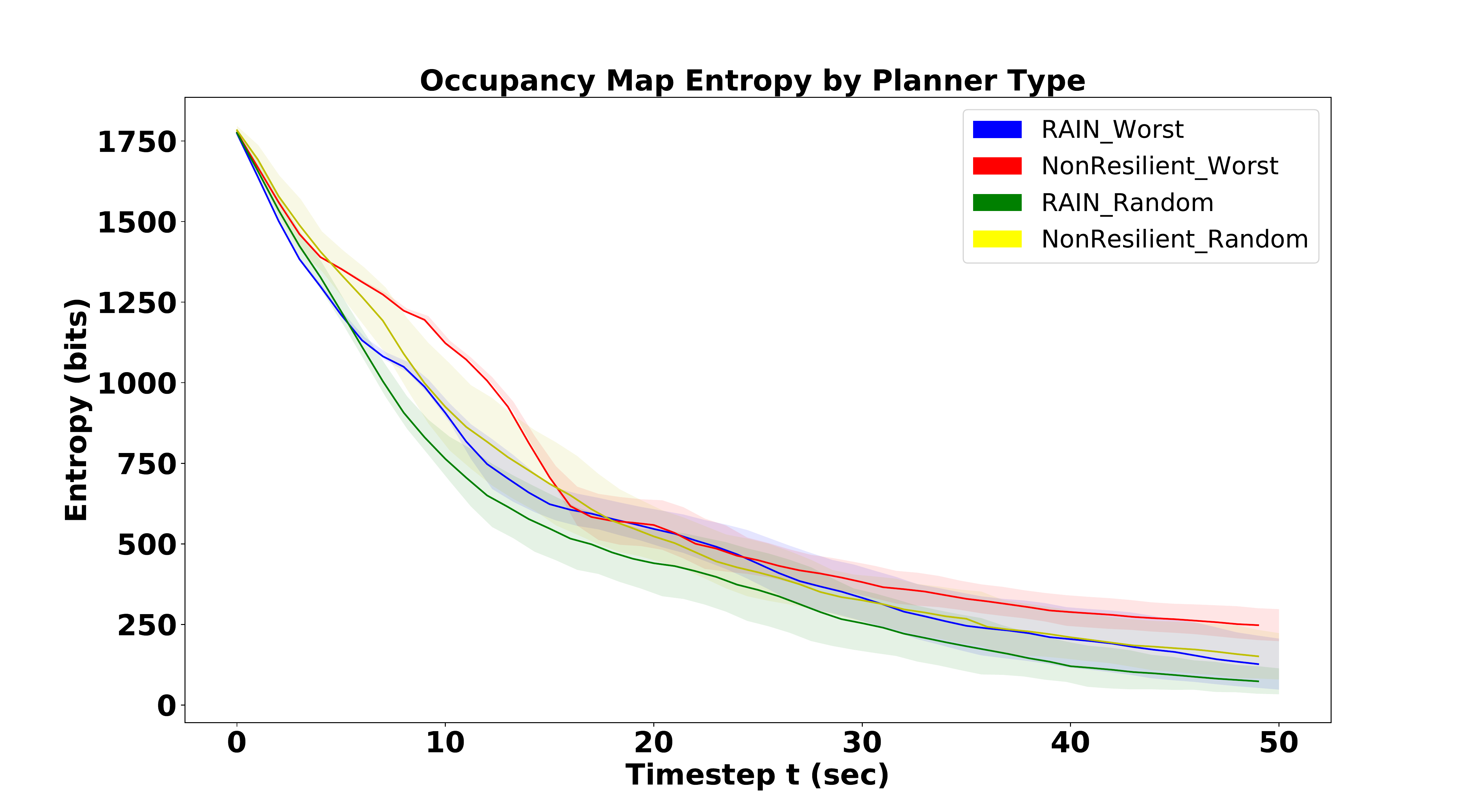}}  
\subfloat[][]{ \hspace{-5mm}\includegraphics[width=280px,height=165px]{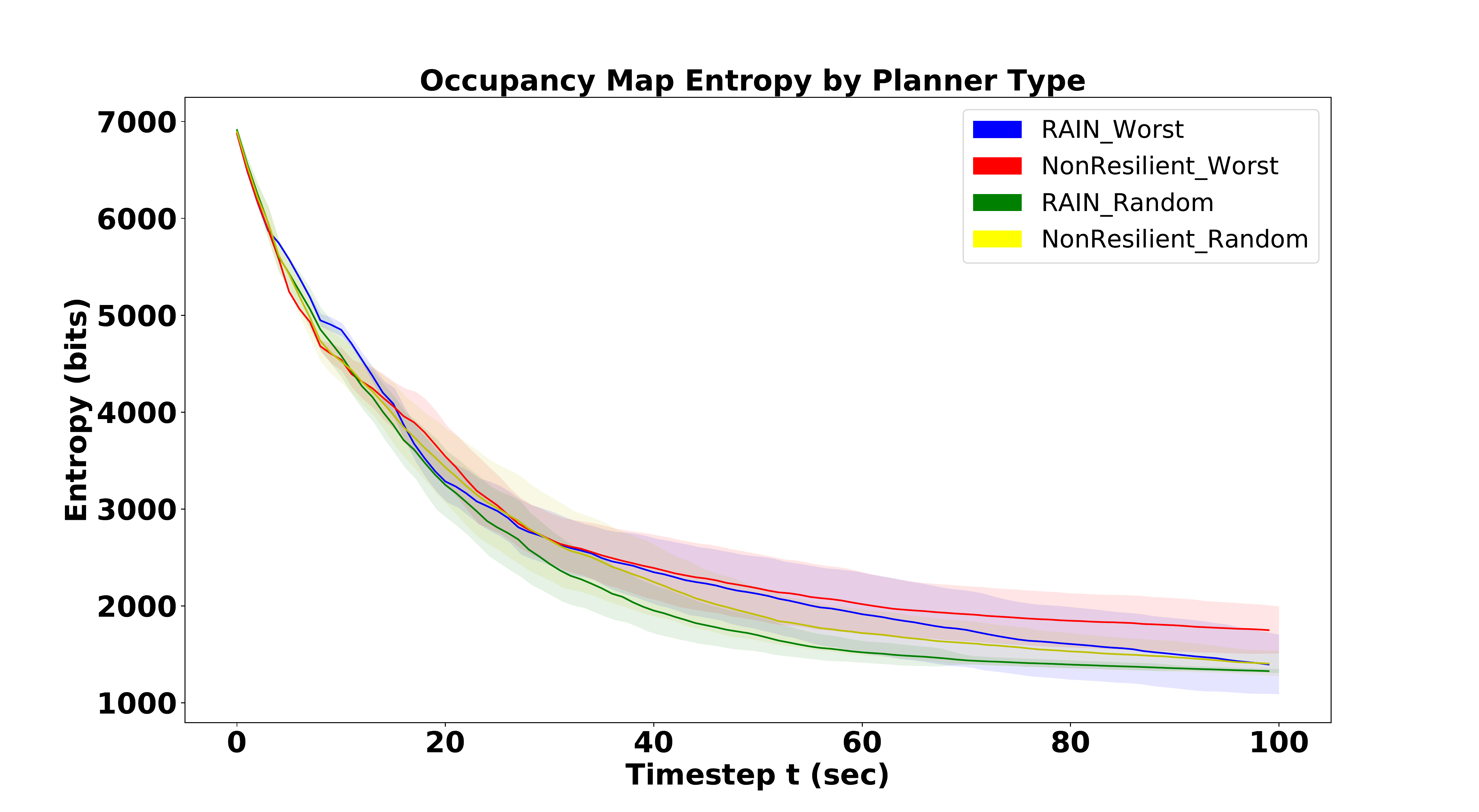}} 
   \vspace{-5mm}
\caption{\textbf{Resilient Occupancy Mapping Results.} 
Comparison of achieved entropy by {\sf \footnotesize RAIN} against coordinate descent (noted as NonResilient in the plots) for increasing time and for two types of attacks, worst-case and random: (left plot) result for the square obstacles map (top row of Fig.~\ref{fig:sim_plots_occ}); (right plot) result for the corridor map (bottom row of Fig.~\ref{fig:sim_plots_occ}). 
}
\label{fig:sim_plots_occ_result1}
\end{figure*}

\subsection{Resilient Occupancy Grid Mapping}\label{subsec:occupancy}
We show how \prone's framework for resilient active information acquisition can be adapted to exploring an environment when the map and objective are defined via occupancy grids. {In this section, we also assess \myalg's sensitivity against non worst-case attacks, in particular, random.}

\myParagraph{Experimental Setup}  {We specify the used (a) \emph{robot dynamics}, (b) \emph{target process}, (c) \emph{sensor model}, (d) \emph{information acquisition objective function}, and (e) \emph{algorithm for solving the optimization problem in \algrobust's line~\ref{line1:opt_1}}:}
\setcounter{paragraph}{0}

\paragraph{{Robot Dynamics}} The robots' dynamics are  as in the multi-target tracking application (Section~\ref{subsec:multi-target}). 

\paragraph{{Target Process}} We define the target process  $y_t$, which we will denote henceforth as $M$ for consistency with common references on occupancy grid mapping \cite{thrun2002probabilistic}, where we also drop the time subscript since the process $y_t$ does not evolve in time. The occupancy grid $M$ is a 2-dimensional grid with $n$ rows and $m$ columns, discretized into cells $M = \{C_1,\ldots, C_{nm}\}$, which are binary variables that are either occupied or free, with some probability. Cell occupancy is assumed to be independent, so that the probability mass function can be factored as $\mathbb{P}(M = m) = \prod_{i=1}^{nm} \mathbb{P}(C_i=c_i)$, where $c_i \in \{0, 1\}$, and where $m \in \{0,1\}^{nm}$ is a particular realization of the map.

\paragraph{{Sensor Model}} We express the sensor model as a series of $B$ beams, such that $z_t = [z_t^1,\ldots,z_t^B]$, where $z_k^b$ is the random  variable of the distance that a beam $b$ travels to intersect an object in the environment. We next define the distribution for a single beam, determined by the true distance $d$ to the first obstacle that the beam intersects:
\begin{align}
    p(z_k^b=z | d) =
    \left\{\begin{array}{ll}
        \mathcal{N}(z-0, \sigma^2),  &  d < z_{\text{min}}; \\
        \mathcal{N}(z-z_{max}, \sigma^2), &   d > z_{\text{max}}; \\
        \mathcal{N}(z-d, \sigma^2), &  \text{otherwise;} 
    \end{array}\right.
\end{align}
$z_{\text{min}}$ and $z_{\text{max}}$ are the minimum and maximum sensing ranges. 

\paragraph{{Information Acquisition Objective Function}} The used information objective function is the Cauchy Schwarz Quadratic Mutual Information (CSQMI), which is shown in the literature to be computationally efficient, as well as, sufficiently accurate for occupancy mapping \cite{charrow2015information}. We denote the CSQMI gathered at time $t$ by $I_{CS}(m;\;z_{\calV, t})$, given the measurements collected at $t$ by the robots in $\calV$. Then, 
\begin{equation*}
\metric_{\calV, 1:\missionlength} \triangleq \frac{1}{{\missionlength}}\sum_{t=1}^{{\missionlength}} I_{CS}(m;\;z_{\calV, t}),
\end{equation*}.

\begin{myremark}[Evaluation of CSQMI] Details on the evaluation of the CSQMI objective are beyond the scope of this paper; we refer the reader to  \cite{charrow2015information}. We note that it relies on a ray-tracing operation for each beam, computed over the current occupancy grid map belief to determine which cells each beam from the LIDAR will observe when the sensor visits a given pose. CSQMI is approximated by assuming the information computed over single beams is additive, but not before pruning approximately independent beams. This removal of approximately independent beams encourages robots to explore areas where their beams will not overlap. In coordinate descent, once a set of prior robots has planned trajectories, future robots must check their beams to see if they are approximately independent from the fixed beams, before their individual beam contributions may be added to the joint CSQMI objective.
\end{myremark}

\paragraph{{Algorithm for Solving Optimization Problem in \algrobust's line~\ref{line1:opt_1}}}
The single-robot motion planning is performed via a full forward search over a short planning horizon $\planninghorizon=4$, since scaling beyond short horizons is challenging in occupancy mapping problems; for details, cf.~\cite{charrow2015information}. 
{{We remark that the performance guarantees do not explicitly hold for the single-robot planner since the measurement model is highly non-linear and the cost function depends on the realization of the measurements, so open-loop planning is not optimal as it is with the Gaussian case~\cite{atanasov2014information}. Nonetheless, approaches similar to what we adopt here have been successfully used for (attack-free) occupancy grid mapping~\cite{charrow2014approximate,corah2019distributed}.}
}

\myParagraph{Simulated Scenarios} 
To evaluate the performance of the resilient occupancy mapping algorithm, we compare the results with different attack types. Namely, we consider an attack model where the attacks on robots may be uniformly random, rather than the worst case attack assumption of the previous section and the algorithm itself. In the experiment, the robots choose trajectories composed of $\planninghorizon=4$ steps of duration $\tau=1$ sec, with motion primitives $\mathcal{U} = \{(v,\omega) : v \in \{0, 1, 2\} m/s,\; \omega \in \{0, \pm 1, \pm 2\} rad/s\}$. The maximum sensor range $r_{sense}$ is $1.5 m$, with a noise standard deviation of $\sigma=.01 m$. The experiment considers a team of six robots, subject to two attack models, described in the following paragraph.  The attacked set of robots  is re-computed at the end of each planning duration. We evaluate the performance on two map environments, which we will refer to as the \emph{square obstacles map} (Fig \ref{fig:sim_plots_occ}, top), and the \emph{corridor map} (Fig \ref{fig:sim_plots_occ}, bottom).  {We use $\missionlength=50$ and $\missionlength=100$ for the squares and corridor map respectively, and $\replanrate=\missionlength$}.

\myParagraph{{Compared Attack Models}} {We test \myalg's ability to be effective even against \textit{non} worst-case failures.  To this end, beyond considering the worst-case attack model prescribed by \prone's problem formulation (cf.~red and blue curves in Fig.~\ref{fig:sim_plots_occ_result1}), we also consider random attacks, chosen with uniformly random assignment among the robots in $\calV$, given the attacks number $\alpha$ (cf.~green and yellow curves Fig.~\ref{fig:sim_plots_occ_result1}).
}

\myParagraph{Results} The results, averaged across {50 Monte-Carlo runs}, are shown in Fig.~\ref{fig:sim_plots_occ_result1}. The plots indicate \myalg always improves performance.  
Specifically, \myalg improves performance both when (i) worst-case attacks are present (cf.~blue and red curves in Fig.~\ref{fig:sim_plots_occ_result1}), and when (ii) random  attacks are present (cf.~green and yellow curves in Fig.~\ref{fig:sim_plots_occ_result1}); in both cases, \myalg attains lesser map entropy against coordinate descent (noted as NonResilient in the plots). 
Both the square obstacles map (left plot in Fig.~\ref{fig:sim_plots_occ_result1}) and the corridor map (right plot in Fig.~\ref{fig:sim_plots_occ_result1}) support this conclusion.
Moreover, Fig.~\ref{fig:sim_plots_occ_result1} supports the intuition that since {\sf \small RAIN} is designed to withstand the worst-case attacks, \myalg's performance will improve when instead only random failures are present (cf.~blue and green curves in Fig.~\ref{fig:sim_plots_occ_result1}). 

\subsection{Resilient Persistent Surveillance}\label{subsec:persistent}
\begin{figure}[t]
   \captionsetup[subfloat]{labelformat=empty}
   \hspace{-0.5mm}\subfloat[][]{ \includegraphics[width=.98\columnwidth]{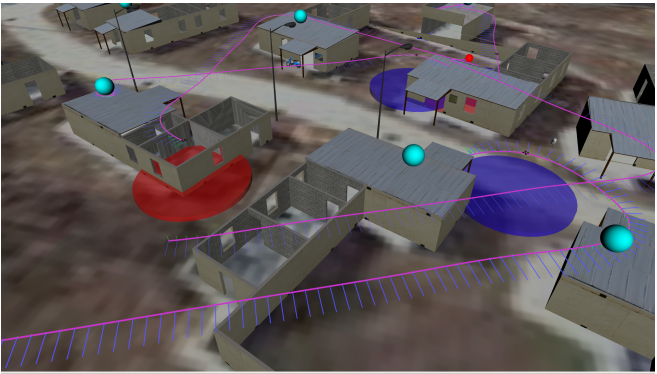}} 
    \vspace{-4mm}
   \caption{
   \textbf{Resilient Persistent Surveillance Scenario.} Camp Lejeune 3D environment. The robots' trajectories are shown in pink. The blue lines along the trajectories indicate the velocity profile generated along the trajectory. The blue discs show the field-of-view of the non-attacked robots.  A red colored field-of-view indicates an attacked robot. The cyan spheres represent the relative uncertainty on the landmarks' locations.  The landmarks are depicted as the red spheres.}
      \label{fig:lejeune}
   \end{figure}

In \emph{Resilient Persistent Surveillance}, the robots' objective is to re-visit a series of static and known landmarks, while the robots are under attack. The landmarks represent points of interest in the environment.  For example, a team of robots may be tasked to monitor the entrances to buildings for intruders~\cite{michael2011persistent}; the task becomes especially interesting as the number of entrances becomes more than the number of robots. 
In this section, we choose the landmarks to be a set of buildings in an outdoor Camp environment (Fig.~\ref{fig:lejeune}).  {We use the simulated scenarios to determine the effect of the replanning rate on \myalg's performance (cf.~Footnote~\ref{foot:replan}).} 

\myParagraph{Experimental Setup} 
\setcounter{paragraph}{0}
The environment used is a 3D environment provided by ARL DCIST (Fig.~\ref{fig:lejeune}). 
It contains a set of outdoor buildings, over which we place landmarks to encourage visitation (one landmark per building). To have the robots (re-)visit the landmarks, we add artificial uncertainty to the location of each landmark by proportionally increasing the uncertainty with time passed since the last observation. The software simulation stack used is based on the Robot Operating System (ROS); the back-end physics are based on Unity. In all experiments, the map is assumed to be known.  Localization is provided by the simulator.
{We next specify  the used (a) \emph{robot dynamics}, (b) \emph{target process}, (c) \emph{sensor model}, and (d) \emph{information acquisition objective function}:}

\paragraph{{Robot Dynamics}} The robot motion model is adapted from the 2D in eq.~\eqref{eq:dynamics} to the following 3D:
\begin{align*}
\scaleMathLine[.88]{
\begin{pmatrix}
x_{t+1}^1 \\ x_{t+1}^2 \\x_{t+1}^3 \\ \theta_{t+1}
\end{pmatrix} = 
\begin{pmatrix}
x_t^1 \\x_t^2 \\ x_t^3 \\ \theta_t 
\end{pmatrix} + 
\begin{pmatrix}
\nu \sinc(\frac{\omega \tau}{2}) \cos (\theta_t + \frac{\omega \tau}{2})\\
\nu \sinc(\frac{\omega \tau}{2}) \sin (\theta_t + \frac{\omega \tau}{2})\\ 
0 \\
\tau \omega
\end{pmatrix}.}
\end{align*}

That is, we assume the quadrotors to fly at a fixed height over the environment ($x^3_{t+1}=x^3_t)$ always).

\paragraph{{Target Process}} 
The targets are assumed static in location, but corrupted with uncertainty that increases over time to encourage (re-)visitation by the robots, according to a noise covariance matrix $qk_{t,m}I_3$, where  $q$ is the rate of uncertainty increase, and $k_{t,m}$ denotes the number of time steps since target $m$ was last visited.
\begin{align*}
y_{t+1,m} = y_{t,m} + w_t, \hspace{3mm} w_t \sim \setN(0, q k_{t,m} I_3).
\end{align*}

\begin{figure}[t]
   \captionsetup[subfloat]{labelformat=empty}
   \centering
   \vspace{-5mm}
   \hspace{-2mm}\subfloat[][]{ \includegraphics[width=0.9999\columnwidth]{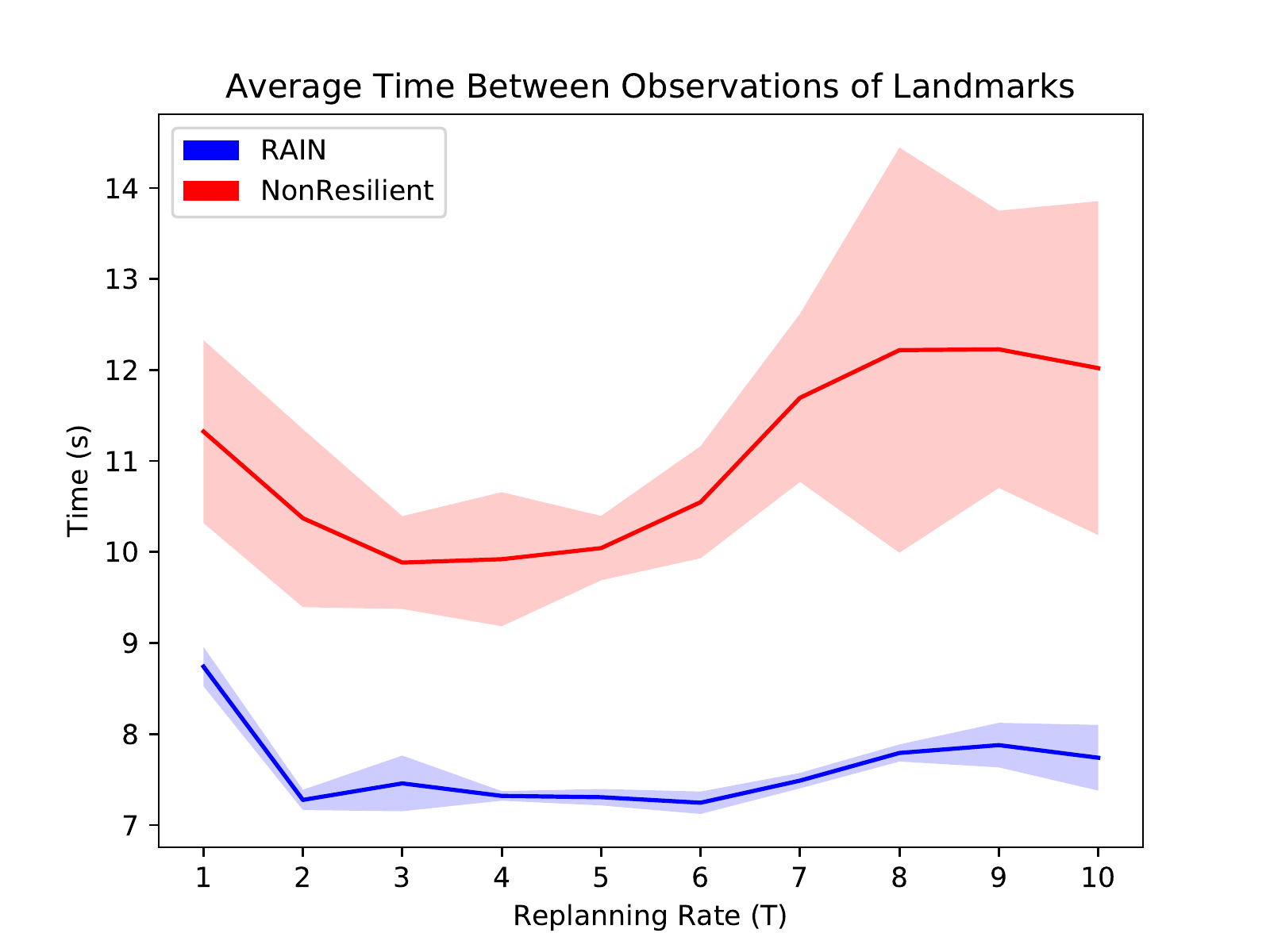}} 
   \vspace{-6mm}
   \caption{\textbf{Resilient Persistent Surveillance Results.} Comparison of average time between consecutive observations of the same landmarks by {\sf \footnotesize RAIN} and coordinate descent (noted as NonResilient in the plot) for increasing replanning values $\replanrate$.}
      \label{fig:surveillance_result}
   \end{figure}

\paragraph{{Sensor Model}} We assume the robots operate a $360^\circ$ field-of-view downward facing sensor.  In particular, we assume a range sensing model that records information as long as the robots are within some radius $r_{sense}$ from a landmark; otherwise, no information is granted to the robot. The range based model for detecting the buildings is as follows:
\begin{align*}
&z_{t,m} = h(x_t,y_{t,m}) + v_t, \hspace{3mm} v_t \sim \setN(0, V(x_t,y_{t,m})); \\
&h(x,y_m) = \begin{bmatrix}r(x,y_m)\end{bmatrix} \triangleq \sqrt{\sum_{i=1}^3 (y^1 - x^1)^2}.
\end{align*}

\paragraph{{Information Acquisition Objective Function}} We use the same information acquisition objective function as in the \emph{Multi-Target Tracking} scenarios (Section~\ref{subsec:multi-target}).

\myParagraph{Simulation Setup} The admissible control inputs are the $\mathcal{U} = \{(\nu,\omega) : \nu \in \{1, 3\} m/s, \omega \in \{0, 1, 2\}\}$.  $r_{sense}$ is $10$ meters, the task duration is $\missionlength=300$ steps, and the planning horizon is $\planninghorizon=10$ steps. Specifically, each timestep has duration $\tau=1s$. The noise parameter is $q=.01$. 

\myParagraph{Performance Metric} We measure \myalg's performance by computing the average number of timesteps that a building goes unobserved for. For example, if a landmark is observed at timestep $k$, and not observed again until timestep $k+l$, we record $l$ as the number of timesteps the landmark was unobserved. Particularly, we average these durations across all targets and timesteps for a given experimental trial.

\myParagraph{Results} The results, averaged across {50 Monte-Carlo runs}, are shown in Fig.~\ref{fig:surveillance_result}. In Fig.~\ref{fig:surveillance_result}, we observe 
even for the highest replanning rate ($\replanrate=1$), \myalg offers a  performance gain of $\simeq 24\%$ in comparison to coordinate descent (noted as NonResilient in Fig.~\ref{fig:surveillance_result}).
The gain increases on average, the lower the replanning rate becomes, as expected (cf.~Footnote~\ref{foot:replan}). 
More broadly, Fig.~\ref{fig:surveillance_result} supports the intuition that a higher replanning rate allows even a non-resilient algorithm, such as coordinate descent, to respond to attacks rapidly, and thus perform well.  Still, in Fig.~\ref{fig:surveillance_result} \myalg dominates coordinate descent across all possible replanning rate values.

\section{Conclusion} \label{sec:con}

We introduced the first receding-horizon framework for \emph{resilient multi-robot path planning} against attacks that disable robots' sensors during information acquisition tasks (cf.~\prone).  
We proposed \emph{Resilient Active Information acquisitioN} (\myalg), a robust and adaptive multi-robot planner against any number of attacks.  
\myalg calls, in an online fashion, \emph{Robust Trajectory Planning} (\algrobust), a subroutine that plans attack-robust control inputs over a look-ahead planning horizon.
{We quantified \algrobust's performance by bounding its suboptimality, using notions of curvature for set function optimization.}  
We demonstrated the necessity for \emph{resilient multi-robot path planning}, as well as \myalg's effectiveness, in information acquisition scenarios of \emph{Multi-Target Tracking}, \emph{Occupancy Grid Mapping}, and \emph{Persistent Surveillance}.  
In all simulations, \myalg was observed to run in real-time, and exhibited superior performance against a state-of-the-art baseline, (non-resilient) \emph{coordinate descent}~\cite{atanasov2015decentralized}.  Across the three scenarios, \myalg's exhibited robustness and superiority even (i) in the presence of a high number of attacks, (ii) against varying models of attacks, and (iii) high replanning rates.  Future work includes extending the proposed framework and algorithms to distributed settings~\cite{corah2019distributed,zhou2019distributed}.



\setcounter{section}{0}
\renewcommand{\thesection}{Appendix \arabic{section}} 
\renewcommand{\thesubsection}{\thesection-\Alph{subsection}}

\section*{Appendices}

In the appendices that follow, we prove Theorem~\ref{th:per_alg_dec_resil_coord_decent} (\ref{app:proof_theorem}) and Proposition~\ref{prop:alg_per_with_coordinate_descent} (\ref{app:proof_proposition}).  To this end, we first present supporting lemmas (\ref{app:supp_lemmas}) and the algorithm \emph{coordinate descent}~\cite{atanasov2015decentralized} (\ref{app:description_coordinate}). We also use the notation:

\myParagraph{Notation}
Consider a finite set $\mathcal{V}$ and a set function $f:2^\mathcal{V}\mapsto \mathbb{R}$. Then, for any set $\mathcal{X}\subseteq \mathcal{V}$ and any set $\mathcal{X}'\subseteq \mathcal{V}$, the symbol $f(\mathcal{X}|\mathcal{X}')$ denotes the marginal value  $f(\mathcal{X}\cup\mathcal{X}')-f(\mathcal{X}')$.  
{We also introduce notation emphasizing that subsets of robots may use different algorithms to compute their control inputs: we let $J(u_{\calA, \;1\mycolon T}^{a}, u_{\calB, \;1\mycolon T}^{b}) \triangleq J_{\calA^{a}\cup \calB^{b}, \;1\mycolon T}$ indicate that the robots in $\calA$ contribute their measurements to  $J$ and their control inputs are chosen with algorithm $a$ (\eg coordinate descent), while robots in $\calB$ also contribute their measurements to $J$ but their inputs are chosen with another algorithm $b$. 
We also occasionally drop the subscript for time indices, since all time indices in the appendices are identical, (namely, $t+1:t+\planninghorizon$). Similarly, when only the set of robots is important, we use the notation $J(\calA) \triangleq\metric_{\calA,\; t+1:t+\planninghorizon}$, for any $\calA\subseteq \calV$. Lastly, the notation $J(\emptyset)$ refers to the information measure evaluated without any measurements from the robot set.}

\section{Preliminary Lemmas}\label{app:supp_lemmas}

The proof of the lemmas is also found in~\cite{tzoumas2017resilient,tzoumas2018resilientSequential}.

\begin{mylemma}\label{lem:non_total_curvature}
Consider a finite set $\mathcal{V}$ and a non-decreasing and submodular set function $f:2^\mathcal{V}\mapsto \mathbb{R}$ such that $f$ is non-negative and $f(\emptyset)=0$. For any $\mathcal{A}\subseteq \mathcal{V}$:
\begin{equation*}
f(\mathcal{A})\geq (1-\kappa_f)\sum_{a \in \mathcal{A}}f(a).
\end{equation*}
\end{mylemma}
\paragraph*{Proof of Lemma~\ref{lem:non_total_curvature}} Let $\mathcal{A}=\{a_1,a_2,\ldots, a_{|{\cal A}|}\}$. We prove Lemma~\ref{lem:curvature} by proving the following two inequalities: 
\begin{align}
f(\mathcal{A})&\geq \sum_{i=1}^{|{\cal A}|} f(a_i|\mathcal{V}\setminus\{a_i\}),\label{ineq:aux_5}\\
\sum_{i=1}^{|{\cal A}|} f(a_i|\mathcal{V}\setminus\{a_i\})&\geq (1-\kappa_f)\sum_{i=1}^{|{\cal A}|} f(a_i)\label{ineq:aux_6}. 
\end{align} 

We begin with the proof of ineq.~\eqref{ineq:aux_5}: 
\begin{align}
f(\mathcal{A})&=f(\mathcal{A}|\emptyset)\label{ineq:aux_9}\\
&\geq f(\mathcal{A}|\mathcal{V}\setminus \mathcal{A})\label{ineq:aux_10}\\
&= \sum_{i=1}^{|{\cal A}|}f(a_i|\mathcal{V}\setminus\{a_i,a_{i+1},\ldots,a_{|{\cal A}|}\})\label{ineq:aux_11}\\
&\geq \sum_{i=1}^{|{\cal A}|}f(a_i|\mathcal{V}\setminus\{a_i\}),\label{ineq:aux_12}
\end{align}
where ineqs.~\eqref{ineq:aux_10} to~\eqref{ineq:aux_12} hold for the following reasons: ineq.~\eqref{ineq:aux_10} is implied by eq.~\eqref{ineq:aux_9} because $f$ is submodular and $\emptyset\subseteq \mathcal{V}\setminus \mathcal{A}$; eq.~\eqref{ineq:aux_11} holds since for any sets $\mathcal{X}\subseteq \mathcal{V}$ and $\mathcal{Y}\subseteq \mathcal{V}$ we have $f(\mathcal{X}|\mathcal{Y})=f(\mathcal{X}\cup \mathcal{Y})-f(\mathcal{Y})$, and also  $\{a_1,a_2,\ldots, a_{|{\cal A}|}\}$ denotes the set $\mathcal{A}$; and ineq.~\eqref{ineq:aux_12} holds since $f$ is submodular and $\mathcal{V}\setminus\{a_i,a_{i+1},\ldots,a_{\mu}\} \subseteq \mathcal{V}\setminus\{a_i\}$.  These observations complete the proof of ineq.~\eqref{ineq:aux_5}.

We now prove ineq.~\eqref{ineq:aux_6} using the Definition~\ref{def:curvature} of $\kappa_f$, as follows: since $\kappa_f=1-\min_{v\in \mathcal{V}}\frac{f(v|\mathcal{V}\setminus\{v\})}{f(v)}$, it is implied that for all elements $v\in \mathcal{V}$ it is $ f(v|\mathcal{V}\setminus\{v\})\geq (1-\kappa_f)f(v)$.  Therefore, adding the latter inequality across all elements $a \in \calA$ completes the proof of ineq.~\eqref{ineq:aux_6}.
\hfill $\blacksquare$

\begin{mylemma}\label{lem:D3}
Consider any finite set $\mathcal{V}$, a non-decreasing and submodular $f:2^\mathcal{V}\mapsto \mathbb{R}$, and non-empty sets $\calY, \calP \subseteq \calV$ such that for all $y \in \calY$ and all $p \in \calP$ $f(y)\geq f(p)$.  Then:
\belowdisplayskip=-12pt\begin{equation*}
f(\calP|\calY)\leq |\calP|f(\calY).
\end{equation*}
\end{mylemma}
\paragraph*{Proof of Lemma~\ref{lem:D3}} Consider any $y \in \calY$ (such an element exists since Lemma~\ref{lem:D3} considers that $\calY$ is non-empty); then, 
\begin{align}
f(\calP|\calY)&= f(\calP\cup\calY)-f(\calY)\label{aux1:1}\\
&\leq f(\calP)+f(\calY)-f(\calY)\label{aux1:2}\\
&= f(\calP)\nonumber\\
&\leq \sum_{p\in\calP}f(p)\label{aux1:4}\\
&\leq |\calP| \max_{p\in\calP} f(p)\nonumber\\
&\leq |\calP|  f(y)\label{aux1:7}\\
&\leq |\calP| f(\calY),\label{aux1:5}
\end{align}
where eq.~\eqref{aux1:1} to ineq.~\eqref{aux1:5} hold for the following reasons: eq.~\eqref{aux1:1} holds since for any sets $\mathcal{X}\subseteq \mathcal{V}$ and $\mathcal{Y}\subseteq \mathcal{V}$, $f(\mathcal{X}|\mathcal{Y})=f(\mathcal{X}\cup \mathcal{Y})-f(\mathcal{Y})$; ineq.~\eqref{aux1:2} holds since $f$ is submodular and, as a result, the submodularity Definition~\ref{def:sub} implies that for any set $\calA\subseteq\calV$ and $\calA'\subseteq\calV$, $f(\calA\cup \calA')\leq f(\calA)+f(\calA')$; ineq.~\eqref{aux1:4} holds for the same reason as ineq.~\eqref{aux1:2}; ineq.~\eqref{aux1:7} holds since for all elements $y \in \calY$ and all elements $p \in \calP$ $f(y)\geq f(p)$; finally, ineq.~\eqref{aux1:5} holds because $f$ is monotone and $y\in\calY$.
\hfill $\blacksquare$

\begin{mylemma}\label{lem:curvature}
Consider a finite set $\mathcal{V}$ and a non-decreasing $\function:2^\mathcal{V}\mapsto \mathbb{R}$ such that $\function$ is non-negative and $\function(\emptyset)=0$. For any set $\mathcal{A}\subseteq \mathcal{V}$ and any set $\mathcal{B}\subseteq \mathcal{V}$ such that $\calA \cap \calB=\emptyset$:
\begin{equation*}
\function(\mathcal{A}\cup \mathcal{B})\geq (1-c_\function)\left(\function(\mathcal{A})+\sum_{b \in \mathcal{B}}\function(b)\right).
\end{equation*}
\end{mylemma}
\paragraph*{Proof of Lemma~\ref{lem:curvature}}
Let $\mathcal{B}=\{b_1, b_2, \ldots, b_{|\mathcal{B}|}\}$. Then, 
\begin{equation}
f(\mathcal{A}\cup \mathcal{B})=\function(\mathcal{A})+\sum_{i=1}^{|\mathcal{B}|}\function(b_i|\mathcal{A}\cup \{b_1, b_2, \ldots, b_{i-1}\}). \label{eq1:lemma_curvature}
\end{equation} 
In addition, Definition~\ref{def:total_curvature} of total curvature implies:
\begin{align}
\function(b_i|\mathcal{A}\cup \{b_1, b_2, \ldots, b_{i-1}\})&\geq (1-c_f)\function(b_i|\emptyset)\nonumber\\
&=(1-c_f)\function(b_i), \label{eq2:lemma_curvature}
\end{align} 
where the latter equation holds since $\function(\emptyset)=0$.
The proof is completed by substituting~\eqref{eq2:lemma_curvature} in~\eqref{eq1:lemma_curvature} and then taking into account that $\function(\mathcal{A})\geq (1-c_f)\function(\mathcal{A})$ since $0\leq c_\function\leq 1$. \hfill $\blacksquare$

\begin{mylemma}\label{lem:subratio}
Consider a finite set $\mathcal{V}$ and a non-decreasing $f:2^\mathcal{V}\mapsto \mathbb{R}$ such that $f$ is non-negative and $f(\emptyset)=0$. For any $\mathcal{A}\subseteq \mathcal{V}$ and any  $\mathcal{B}\subseteq \mathcal{V}$ such that $\mathcal{A}\setminus\mathcal{B}\neq \emptyset$:
\begin{equation*}
f(\mathcal{A})+(1-c_f) f(\mathcal{B})\geq (1-c_f) f(\mathcal{A}\cup \mathcal{B})+f(\mathcal{A}\cap \mathcal{B}).
\end{equation*}
\end{mylemma}
\paragraph*{Proof of Lemma~\ref{lem:subratio}}
Let $\mathcal{A}\setminus\mathcal{B}=\{i_1,i_2,\ldots, i_r\}$, where $r=|\mathcal{A}-\mathcal{B}|$. From Definition~\ref{def:total_curvature} of total curvature $c_f$, for any $i=1,2, \ldots, r$, it is  $f(i_j|\mathcal{A} \cap \mathcal{B} \cup \{i_1, i_2, \ldots, i_{j-1}\})\geq (1-c_f) f(i_j|\mathcal{B} \cup \{i_1, i_2, \ldots, i_{j-1}\})$. Summing these $r$ inequalities,
$$f(\mathcal{A})-f(\mathcal{A}\cap \mathcal{B})\geq (1-c_f) \left(f(\mathcal{A}\cup \mathcal{B})-f(\mathcal{B})\right),$$
which implies the lemma. \hfill $\blacksquare$

\begin{mycorollary}\label{cor:ineq_from_lemmata}
Consider a finite set $\mathcal{V}$ and a non-decreasing $f:2^\mathcal{V}\mapsto \mathbb{R}$ such that $f$ is non-negative and $f(\emptyset)=0$. For any $\mathcal{A}\subseteq \mathcal{V}$ and any $\mathcal{B}\subseteq \mathcal{V}$ such that $\mathcal{A}\cap\mathcal{B}=\emptyset$:
\begin{equation*}
f(\mathcal{A})+\sum_{b \in \mathcal{B}}f(b) \geq (1-c_f)  f(\mathcal{A}\cup \mathcal{B}).
\end{equation*}
\end{mycorollary}
\paragraph*{Proof of Corollary~\ref{cor:ineq_from_lemmata}}
 Let $\mathcal{B}=\{b_1,b_2,\ldots,b_{|\mathcal{B}|}\}$. 
\begin{align}
f(\mathcal{A})+\sum_{i=1}^{|\mathcal{B}|}f(b_i) &\geq (1-c_f) f(\mathcal{A})+\sum_{i=1}^{|\mathcal{B}|}f(b_i))\label{ineq:cor_aux1} \\
& \geq (1-c_f) f(\mathcal{A}\cup \{b_1\})+\sum_{i=2}^{|\mathcal{B}|}f(b_i)\nonumber\\
& \geq (1-c_f) f(\mathcal{A}\cup \{b_1,b_2\})+\sum_{i=3}^{|\mathcal{B}|}f(b_i)\nonumber\\
& \;\;\vdots \nonumber\\
& \geq (1-c_f) f(\mathcal{A}\cup \mathcal{B}),\nonumber
\end{align}
where~\eqref{ineq:cor_aux1} holds since $0\leq c_f\leq 1$, and the rest due to Lemma~\ref{lem:subratio} since $\mathcal{A}\cap\mathcal{B}=\emptyset$ implies $\mathcal{A}\setminus \{b_1\}\neq \emptyset$, $\mathcal{A}\cup \{b_1\}\setminus \{b_2\}\neq \emptyset$, $\ldots$, $\mathcal{A}\cup \{b_1,b_2,\ldots, b_{|\mathcal{B}|-1}\}\setminus \{b_{|\mathcal{B}|}\}\neq \emptyset$. 

\hfill $\blacksquare$

\section{Coordinate Descent}\label{app:description_coordinate}

We describe \textit{coordinate descent}~\cite[Section~IV]{atanasov2015decentralized}, and generalize the proof in \cite{atanasov2015decentralized} that \textit{coordinate descent} guarantees an approximation performance up to a multiplicative factor $1/2$ the optimal when the information objective function is the mutual information.  In particular, we extend the proof to \textit{any} non-decreasing and possibly submodular information objective function; the result will support the proof of Proposition~\ref{prop:alg_per_with_coordinate_descent}.

The algorithm coordinate descent  works as follows: consider an \textit{arbitrary} ordering of the robots in $\calV$, such that $\calV \equiv \{1,2,\ldots, n\}$, and suppose that robot $1$ first chooses its controls, without considering the other robots; in other words, robot $1$ solves the single robot version of Problem~\ref{pr:rtp}, \ie $J_{\{1\}, t+1:t+\planninghorizon} := J(u_1)$, to obtain controls $u_{\{1\}}$ such that:

\begin{equation} \label{eq:descent}
\begin{aligned}
u^{cd}\at{1}{t+1}{t+\planninghorizon} = \arg \optmshortSingleRobot \metric(u\at{i}{t+1}{t+\planninghorizon})
\end{aligned}
\end{equation}

 Afterwards, robot $1$ communicates its chosen control sequence to robot 2, and robot 2, given the control sequence of robot~1.computes its control input as follows, assuming the control inputs for robot~1 are fixed:
\begin{equation}
\begin{aligned}
u^{cd}\at{2}{t+1}{t+\planninghorizon} = \arg \optmshortSingleRobot \metric(u^{cd}_1, u\at{i}{t+1}{t+\planninghorizon})
\end{aligned}
\end{equation}

This continues such that robot $i$ solves a single robot problem, given the control inputs from the robots $1,2,\ldots,i-1$:
\begin{equation}
\begin{aligned}
u^{cd}\at{i}{t+1}{t+\planninghorizon} = \arg \optmshortSingleRobot \!\!\metric(u^{cd}_{1:i-1}, u\at{i}{t+1}{t+\planninghorizon})
\end{aligned}
\end{equation}

Notably, if we let $u_i^{*}$ be the control inputs for the $i$-th robot resulting from the optimal solution to the $n$ robot problem, then from the coordinate descent algorithm, we have:
\begin{equation}\label{cd_policy}
\begin{aligned}
J(u^{cd}_{1:i-1}, u^\star_i) \leq J(u^{cd}_{1:i})
\end{aligned}
\end{equation}


\begin{mylemma}[Approximation performance of coordinate descent]\label{lem:generalized_cd} 
Consider a set of robots $\calV$, and an instance of problem $\prtwo$. Denote the optimal control inputs for problem $\prtwo$, across all robots and all times, by $u^\star_{\calV, t+1\;:\;t+\planninghorizon}$. The coordinate descent algorithm returns control inputs $u^{cd}_{\calV,\; t+1\;:\;t+\planninghorizon}$, across all robots and all times, such that:
\begin{itemize}
    \item if the objective function $\metric$ is non-decreasing submodular in the active robot set, and (without loss of generality) $\metric$ is non-negative and $\metric(\emptyset)=0$, then: 
\begin{equation}\label{ineq:bound_cd_1}
\frac{J(u_{1:n}^{cd})}{J(u_{1:n}^\star)} \geq \frac{1}{2}.
\end{equation}
\item If the objective function $\metric$ is non-decreasing in the active robot set, and (without loss of generality) $\metric$ is non-negative and $\metric(\emptyset)=0$, then: 
\begin{equation}\label{ineq:bound_cd_2}
\frac{J(u_{1:n}^{cd})}{J(u_{1:n}^\star)} \geq \frac{1-c_\metric}{2}.
\end{equation}
\end{itemize}
\end{mylemma}

\paragraph*{Proof of Lemma~\ref{lem:generalized_cd}}
\begin{itemize}
\item if the objective function $\metric$ is non-decreasing and submodular in the active robot set, and (without loss of generality) $\metric$ is non-negative and $\metric(\emptyset)=0$, then:
\begin{alignat}{3}
\label{cd:ineq_mono}J(u^\star_{1:n}) &\leq J(u^\star_{1:n}) + \sum_{i=1}^n [J(u^{cd}_{1:i}, u^\star_{i+1:n}) \\ 
&\hspace{25mm}-J(u^{cd}_{1:i-1}, u^\star_{i+1:n})] \nonumber \\
&\label{cd:eq_arrange} = J(u^{cd}_{1:n}) + \sum_{i=1}^{n}[ J(u^{cd}_{1:i-1}, u^\star_{i:n})  \\
&\hspace{25mm} -J(u^{cd}_{1:i-1}, u^\star_{i+1:n})]\nonumber \\
 &= J(u^{cd}_{1:n}) + \sum_{i=1}^n J(u^\star_i | \{u^{cd}_{1:i-1}, u^\star_{i+1,n} \})\label{cd:eq_marginal} \\
&\leq J(u^{cd}_{1:n}) + \sum_{i=1}^n J(u^\star_i | u^{cd}_{1:i-1}) \label{cd:ineq_submod}\\
&\leq J(u^{cd}_{1:n}) + \sum_{i=1}^n J(u^{cd}_i | u^{cd}_{1:i-1})\label{cd:ineq_cd} 
\end{alignat}

\begin{alignat}{3}
&= J(u^{cd}_{1:n}) + J(u^{cd}_{1:n}) \label{cd:eq_marginal_sum} \\
&\leq 2J(u^{cd}_{1:n}), \label{cd:eq_marginal_sum_final}
\end{alignat}
where ineq.~(\ref{cd:ineq_mono}) holds due to monotonicity of $J$; eq.~\ref{cd:eq_arrange}) is a shift in indexes of the first term in the sum; eq.~(\ref{cd:eq_marginal}) is an expression of the sum as a sum of marginal gains; ineq.~(\ref{cd:ineq_submod}) holds due to submodularity; ineq.~(\ref{cd:ineq_cd}) holds by the coordinate-descent policy (per~eq.~\eqref{cd_policy}); eq.~(\ref{cd:eq_marginal_sum}) holds due to the definition of the marginal gain symbol $J(u^\star_i | u^{cd}_{1:i-1})$ (for any $i=1,2,\ldots,n$) as $J(u^\star_i , u^{cd}_{1:i-1})-J(u^{cd}_{1:i-1})$; finally, a re-arrangement of the terms in eq.~\eqref{cd:eq_marginal_sum_final} gives $J(u^{cd}_{1:n})/J(u^{*}_{1:n}) \geq 1/2$.

\item If $\metric$ is non-decreasing in the active robot set, and (without loss of generality) $\metric$ is non-negative and $\metric(\emptyset)=0$, then multiplying both sides of eq.~\eqref{cd:eq_marginal} (which holds for any non-decreasing $\metric$) with $(1-c_\metric)$, we have:
\begin{alignat}{3}
 (1-&c_\metric)J(u^\star_{1:n}) \nonumber\\
 &= (1-c_\metric)J(u^{cd}_{1:n}) + \nonumber\\
 &\hspace{2cm}(1-c_\metric)\sum_{i=1}^n J(u^\star_i | \{u^{cd}_{1:i-1}, u^\star_{i+1,n} \})\nonumber\\
 &\leq J(u^{cd}_{1:n}) + (1-c_\metric)\sum_{i=1}^n J(u^\star_i | \{u^{cd}_{1:i-1}, u^\star_{i+1,n} \})\label{cd2:eq_marginal2} \\
&\leq J(u^{cd}_{1:n}) + \sum_{i=1}^n J(u^\star_i | u^{cd}_{1:i-1}) \label{cd2:ineq_submod}\\
&\leq J(u^{cd}_{1:n}) + \sum_{i=1}^n J(u^{cd}_i | u^{cd}_{1:i-1})\label{cd2:ineq_cd} \\
&= J(u^{cd}_{1:n}) + J(u^{cd}_{1:n}) \label{cd2:eq_marginal_sum} \\
&\leq 2J(u^{cd}_{1:n}),
\end{alignat}
where, ineq.~\eqref{cd2:eq_marginal2} holds since $0\leq c_\metric\leq 1$; ineq.~(\ref{cd2:ineq_submod}) holds since $J$ is non-decreasing in the set of active robots, and Definition~\ref{def:total_curvature} of total curvature implies that for any non-decreasing  set function $g:2^\calV\mapsto\mathbb{R}$, for any element $\elem\in\calV$, and for any set $\calA,\calB\subseteq \calV\setminus \{\elem\}$: 
 \begin{equation}\label{eq:ineq_total_curvature}
 (1-c_g) g(\elem|\calB)\leq g(\{\elem\}|\calA);
  \end{equation}
ineq.~(\ref{cd2:ineq_cd}) holds by the coordinate-descent algorithm; eq.~(\ref{cd2:eq_marginal_sum}) holds due to the definition of the marginal gain symbol $J(u^\star_i | u^{cd}_{1:i-1})$ (for any $i=1,2,\ldots,n$) as $J(u^\star_i , u^{cd}_{1:i-1})-J(u^{cd}_{1:i-1})$; finally, a re-arrangement of terms gives $J(u^{cd}_{1:n})/J(u^{*}_{1:n}) \geq (1-c_\metric)/2$.  \hfill $\blacksquare$
\end{itemize}

\section{Proof of Theorem~\ref{th:per_alg_dec_resil_coord_decent}}\label{app:proof_theorem}

We first prove Theorem~\ref{th:per_alg_dec_resil_coord_decent}'s part 1 (approximation performance), and then, Theorem~\ref{th:per_alg_dec_resil_coord_decent}'s part 2 (running time). 

\subsection{Proof of Theorem~\ref{th:per_alg_dec_resil_coord_decent}'s Part 1 (Approximation Performance)}

The proof follows the steps of the proof of~\cite[Theorem~1]{tzoumas2017resilient} and~\cite[Theorem~1]{tzoumas2018resilientSequential}.  
We first prove eq.~\eqref{ineq:bound_sub}, then, eq.~\eqref{ineq:bound_non_sub}.

To the above ends, we use the following notation (along with the notation introduced in Theorem~\ref{th:per_alg_dec_resil_coord_decent} and in Appendix~A): given that using Algorithm~\ref{alg:dec_resil_coord_decent} the robots in $\calV$ select control inputs $u_{\calV, t+1:t+\planninghorizon}$, then, for notational simplicity:
\begin{itemize}
\item let $\calA^\star\triangleq\calA^\star(u_{\calV,\; t+1:t+\planninghorizon})$;
\item let $\calL^+\triangleq \calL\setminus \calA^\star$, \ie $\calS_1$ be the remaining robots in $\calL$ after the removal of the robots in $\calA^\star$;
\item let $(\calV\setminus\calL)^+\triangleq (\calV\setminus\calL)\setminus \calA^\star$, \ie $\calS_2$ be the remaining robots in $\calV\setminus\calL$ after the removal of the robots in $\calA^\star$.
\end{itemize}

\paragraph*{Proof of ineq.~\eqref{ineq:bound_sub}}
The proof follows the steps of the proof of~\cite[Theorem~1]{tzoumas2017resilient}.
Consider that the objective function $\metric$ is non-decreasing and submodular in the active robot set, such that (without loss of generality) $\metric$ is non-negative and $\metric(\emptyset)=0$.  We first prove the part $1-\kappa_\metric$ of the bound in the right-hand-side of ineq.~\eqref{ineq:bound_sub}, and then, the part $h(|\calV|,\attack)$ of the bound in the right-hand-side of ineq.~\eqref{ineq:bound_sub}.

To prove the part $1-\kappa_\metric$ of the bound in the right-hand-side of ineq.~\eqref{ineq:bound_sub}, we follow the steps of the proof of~\cite[Theorem~1]{tzoumas2017resilient},
and make the following observations:
\begin{align}
& \metric(\mathcal{V}\setminus\mathcal{A}^\star) \nonumber\\
&\;\;\;= \metric(\calL^+\cup (\calV\setminus\calL)^+)\label{ineq:aux_14}\\
&\;\;\;\geq (1-\kappa_\metric)\sum_{v \in \calL^+\cup (\calV\setminus\calL)^+}\metric(v)\label{ineq:aux_15}\\
&\;\;\;\geq (1-\kappa_\metric)\left(\sum_{v \in (\mathcal{V}\setminus \mathcal{L})\setminus (\calV\setminus\calL)^+}\metric(v)+\sum_{v \in (\calV\setminus\calL)^+}\metric(v)\right)\label{ineq:aux_16}\\
&\;\;\;\geq (1-\kappa_\metric)\metric\{[(\mathcal{V}\setminus \mathcal{L})\setminus (\calV\setminus\calL)^+]\cup (\calV\setminus\calL)^+\}\label{ineq:aux_17}\\
&\;\;\;= (1-\kappa_\metric)\metric(\calV\setminus\calL)\label{ineq:aux_18},
\end{align}
where eq.~\eqref{ineq:aux_14} to~\eqref{ineq:aux_18} hold for the following reasons: eq.~\eqref{ineq:aux_14} follows from the definitions of the sets~$\mathcal{L}^+$ and $(\calV\setminus\calL)^+$; ineq.~\eqref{ineq:aux_15} follows from ineq.~\eqref{ineq:aux_14} due to Lemma~\ref{lem:non_total_curvature}; ineq.~\eqref{ineq:aux_16} follows from ineq.~\eqref{ineq:aux_15} because for all elements $v \in \mathcal{L}^+$ and all elements  $v' \in (\mathcal{V}\setminus \mathcal{L})\setminus (\calV\setminus\calL)^+$ we have $\metric(v)\geq \metric(v')$ (note that due to the definitions of the sets~$\mathcal{L}^+$ and $(\calV\setminus\calL)^+$, $|\mathcal{L}^+|=|(\mathcal{V}\setminus \mathcal{L})\setminus (\calV\setminus\calL)^+|$, that is, the number of non-removed elements in $\calL$ is equal to the number of removed elements in $\calV\setminus \calL$);  finally, ineq.~\eqref{ineq:aux_17} follows from ineq.~\eqref{ineq:aux_16} because the set function $\metric$ is submodular and, as~a result, the~submodularity Definition~\ref{def:sub} implies that for any sets $\mathcal{S}\subseteq \mathcal{V}$ and $\mathcal{S}'\subseteq \mathcal{V}$, $\metric(\mathcal{S})+\metric(\mathcal{S}')\geq \metric(\mathcal{S}\cup \mathcal{S}')$~\cite[Proposition 2.1]{nemhauser78analysis}. We now complete the proof of the part $1-\kappa_\metric$ of the bound in the right-hand-side of ineq.~\eqref{ineq:bound_sub} by proving that in ineq.~\eqref{ineq:aux_18}:
\begin{equation}\label{ineq:main_ineq}
\metric(\calV\setminus\calL)\geq\metric^\star\!\!,
\end{equation}
when the robots in $\calV$ optimally solve the problems in Algorithm~\ref{alg:dec_resil_coord_decent}'s  step~\ref{line1:step_4}, per the statement of Theorem~\ref{th:per_alg_dec_resil_coord_decent}.  In particular, if for any active robot set $\calR\subseteq \calV$, we let $\bar{u}_\calR \triangleq \{\bar{u}_{i,t'}:~~\bar{u}_{i,t'}\in \calU_{i,t'},~~i\in\calR,~~t' =t+1,\ldots,t+\planninghorizon\}$ denote a collection of control inputs to the robots in $\calR$, then:
\begin{align}
\metric(\calV\setminus \calL)&\equiv\max_{\scriptsize\begin{array}{c}
\bar{u}_{i,t} \in \mathcal{U}_{i,t},i\in\calV,\\
 t'=t+1:t+\planninghorizon
\end{array}}\metric(\bar{u}_{\calV\setminus\calL, t+1:t+\planninghorizon})\label{ineq:main_ineq_aux1}\\
&\geq \min_{\scriptsize\begin{array}{c}\bar{\calL}\subseteq \calV,\\
|\bar{\calL}|\leq \attack
\end{array}}\max_{\scriptsize\begin{array}{c}
\bar{u}_{i,t} \in \mathcal{U}_{i,t},i\in\calV,\\
 t'=t+1:t+\planninghorizon
\end{array}}
\metric(\bar{u}_{\calV\setminus\bar{\calL}, t+1:t+\planninghorizon})\label{ineq:main_ineq_aux2}
\end{align}
\begin{align}
&\geq \max_{\scriptsize\begin{array}{c}
\bar{u}_{i,t} \in \mathcal{U}_{i,t},i\in\calV,\\
 t=t+1:t+\planninghorizon
\end{array}}\min_{\scriptsize\begin{array}{c}\bar{\calL}\subseteq \calV,\\
|\bar{\calL}|\leq \attack
\end{array}}\metric(\bar{u}_{\calV\setminus\bar{\calL}, t+1:t+\planninghorizon})\label{ineq:main_ineq_aux3}\\
&\equiv \metric^\star\!\!,\label{ineq:main_ineq_aux4}
\end{align}
where~\eqref{ineq:main_ineq_aux1}-\eqref{ineq:main_ineq_aux4} hold true since: the equivalence in eq.~\eqref{ineq:main_ineq_aux1} holds since the robots in $\calV$ solve optimally the problems in Algorithm~\ref{alg:dec_resil_coord_decent}'s  step~\ref{line1:step_4}, per the statement of Theorem~\ref{th:per_alg_dec_resil_coord_decent};~\eqref{ineq:main_ineq_aux2} holds since we minimize over the set $\calL$;~\eqref{ineq:main_ineq_aux3} holds because for any set $\hat{\calL}\subseteq \calV$ and any control inputs $\hat{u}_{\calR, t+1:t+\planninghorizon} \triangleq \{\hat{u}_{i,t}:~~\hat{u}_{i,t}\in \calU_{i,t},~~i\in\calR,~~t' =t+1,\ldots,t+\planninghorizon\}$:
\begin{equation*}
\!\!\!{\max_{\scriptsize\begin{array}{c}
\bar{u}_{i,t} \in \mathcal{U}_{i,t},i\in\calV,\\
 t'=t+1:t+\planninghorizon
\end{array}}\!\!\!\!\metric(\bar{u}_{\calV\setminus\bar{\calL}, t+1:t+\planninghorizon}) \geq \metric(\hat{u}_{\calV\setminus\hat{\calL}, t+1:t+\planninghorizon}),}
\end{equation*}
which implies 
\begin{align*}
&\min_{\scriptsize\begin{array}{c}\bar{\calL}\subseteq \calV,\\
|\bar{\calL}|\leq \attack
\end{array}}\max_{\scriptsize\begin{array}{c}
\bar{u}_{i,t} \in \mathcal{U}_{i,t},i\in\calV,\\
 t'=t+1:t+\planninghorizon
\end{array}}\metric(\bar{u}_{\calV\setminus\bar{\calL}, t+1:t+\planninghorizon})\geq \\
\vspace*{-2mm}
&\hspace{3cm}\min_{\scriptsize\begin{array}{c}\bar{\calL}\subseteq \calV,\\
|\bar{\calL}|\leq \attack
\end{array}}\metric(\hat{u}_{\calV\setminus\bar{\calL}, t+1:t+\planninghorizon}) \Rightarrow\\
\vspace{4mm}
&\min_{\scriptsize\begin{array}{c}\bar{\calL}\subseteq \calV,\\
|\bar{\calL}|\leq \attack
\end{array}}\max_{\scriptsize\begin{array}{c}
\bar{u}_{i,t} \in \mathcal{U}_{i,t},i\in\calV,\\
 t'=t+1:t+\planninghorizon
\end{array}}\metric(\bar{u}_{\calV\setminus\bar{\calL}, t+1:t+\planninghorizon})\geq \\
&\hspace{1cm}\max_{\scriptsize\begin{array}{c}
\bar{u}_{i,t} \in \mathcal{U}_{i,t},i\in\calV,\\
 t'=t+1:t+\planninghorizon
\end{array}}\min_{\scriptsize\begin{array}{c}\bar{\calL}\subseteq \calV,\\
|\bar{\calL}|\leq \attack
\end{array}}\metric(\bar{u}_{\calV\setminus\bar{\calL}, t+1:t+\planninghorizon}),
\end{align*}
where the last one is eq.~\eqref{ineq:main_ineq_aux3}; 
finally, the equivalence in eq.~\eqref{ineq:main_ineq_aux4} holds since $\metric^\star$ (per the statement of Theorem~\ref{th:per_alg_dec_resil_coord_decent}) denotes the optimal value to Problem~\ref{pr:rtp}. 
Overall, we proved that ineq.~\eqref{ineq:main_ineq_aux4} proves ineq.~\eqref{ineq:main_ineq}; and, now, the combination of ineq.~\eqref{ineq:aux_18} and ineq.~\eqref{ineq:main_ineq} proves the part $1-\kappa_\metric$ of the bound in the right-hand-side of ineq.~\eqref{ineq:bound_sub}.

\begin{figure}[t]
\def \setAone{ (0.25,0) circle (1cm) }
\def \setBone{ (.75,0) circle (0.4cm)}
\def \setAtwo{ (2.75,0) circle (1cm) }
\def \setBtwo{ (2.9,0) circle (0.4cm)}
\def \myrectangle{ (-1.5, -1.5) rectangle (4, 1.5) }
\begin{center}
\begin{tikzpicture}
\draw \myrectangle node[below left]{$\mathcal{V}$};
\draw \setAone node[left]{$\mathcal{L}$};
\draw \setBone node[]{$\mathcal{A}_1^\star$};
\draw \setBtwo node[]{$\mathcal{A}_2^\star$};
\end{tikzpicture}
\end{center}
\caption{\small Venn diagram, where the set $\mathcal{L}$ is the robot set defined in step~\ref{line1:step_2} of Algorithm~\ref{alg:dec_resil_coord_decent}, and the set  $\mathcal{A}_1^\star$ and the set $\mathcal{A}_2^\star$ are such that  $\mathcal{A}_1^\star=\mathcal{A}^\star\cap\mathcal{L}$, and $\mathcal{A}_2^\star=\mathcal{A}^\star\cap(\calV\setminus\mathcal{L})$ (observe that these definitions imply $\mathcal{A}_1^\star\cap \mathcal{A}_2^\star=\emptyset$ and $\mathcal{A}^\star=\mathcal{A}_1^\star\cup \mathcal{A}_2^\star$).  
}\label{fig:venn_diagram_for_proof}
\end{figure}
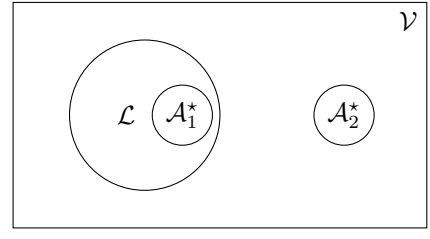

We finally prove the part $1/(1+\alpha)$ of the bound in the right-hand-side of ineq.~\eqref{ineq:bound_sub}, and complete this way the proof of Theorem~\ref{th:per_alg_dec_resil_coord_decent}.  
To this end, we follow the steps of the proof of~\cite[Theorem~1]{tzoumas2017resilient}, and use the notation introduced in Fig.~\ref{fig:venn_diagram_for_proof}, along with the following notation:
\begin{align}
	\eta \triangleq\frac{\metric(\mathcal{A}_2^\star|\calV\setminus \calA^\star)}{\metric(\mathcal{V}\setminus \calL)}
\end{align}
Later in this proof, we prove $0\leq \eta\leq 1$.  We first observe that:
\begin{equation}\label{ineq:aux_1}
\metric(\calV\setminus \calA^\star)\geq\max\{\metric(\calV\setminus \calA^\star),\metric(\calL^+)\};
\end{equation}
in the following paragraphs, we prove the three inequalities:
\begin{align}
\metric(\calV\setminus \calA^\star)&\geq(1-\eta)\metric(\mathcal{V}\setminus \calL)\label{ineq:aux_2},\\
\metric(\calL^+)&\geq \eta \frac{1}{\attack}\metric(\mathcal{V}\setminus \calL),\label{ineq:aux_3}\\
\max\{(1-\eta),\eta\frac{1}{\attack}\}&\geq \frac{1}{\attack+1}.\label{ineq:aux_4}
\end{align}
Then, if we substitute ineq.~\eqref{ineq:aux_2}, ineq.~\eqref{ineq:aux_3} and ineq.~\eqref{ineq:aux_4} to ineq.~\eqref{ineq:aux_1}, and take into account that $\metric(\mathcal{V}\setminus \calL)\geq 0$, then:
\begin{equation*}
\metric(\calV\setminus \calA^\star)\geq \frac{1}{\attack+1}\metric(\mathcal{V}\setminus \calL),
\end{equation*}
which implies the part $1/(1+\alpha)$ of the bound in the right-hand-side of ineq.~\eqref{ineq:bound_sub}, after taking into account ineq.~\eqref{ineq:main_ineq}.

We next complete the proof of the part $1/(1+\alpha)$ of the bound in the right-hand-side of ineq.~\eqref{ineq:bound_sub} by proving $0\leq \eta\leq 1$, ineq.~\eqref{ineq:aux_2}, ineq.~\eqref{ineq:aux_3}, and ineq.~\eqref{ineq:aux_4}.

\setcounter{paragraph}{0}
\paragraph{Proof of ineq.~$0\leq \eta\leq 1$} We first prove $\eta\geq 0$, and then $\eta\leq 1$:~$\eta\geq 0$, since $\eta\equiv\metric(\mathcal{A}_2^\star|\calV\setminus \calA^\star)/\metric(\mathcal{V}\setminus\calL)$, and $\metric$ is non-negative; and~$\eta\leq 1$, since $\metric(\mathcal{V}\setminus\calL)\geq \metric(\mathcal{A}^\star_2)$, due to monotonicity of $\metric$ and that $\mathcal{A}^\star_2 \subseteq \mathcal{V}\setminus\calL$, and $\metric(\mathcal{A}^\star_2)\geq \metric(\mathcal{A}_2^\star|\calV\setminus \calA^\star)$, due to submodularity of $\metric$ and that $\emptyset \subseteq \calV\setminus \calA^\star$. 


\paragraph{Proof of ineq.~\eqref{ineq:aux_2}}  We complete the proof of ineq.~\eqref{ineq:aux_2} in two steps.  First, it can be verified that:
\begin{align}\label{eq:aux_1}
& f(\calV\setminus \calA^\star)=f(\mathcal{V}\setminus\calL)-\nonumber\\ & \metric(\mathcal{A}^\star_2|\calV\setminus \calA^\star)+\metric(\mathcal{L}|\mathcal{V}\setminus\calL)-\metric(\mathcal{A}^\star_1|\mathcal{V}\setminus\mathcal{A}^\star_1),
\end{align}
since for any sets $\mathcal{X}\subseteq \mathcal{V}$ and $\mathcal{Y}\subseteq \mathcal{V}$, $\metric(\mathcal{X}|\mathcal{Y})\equiv\metric(\mathcal{X}\cup \mathcal{Y})-\metric(\mathcal{Y})$. Second, eq.~\eqref{eq:aux_1} implies ineq.~\eqref{ineq:aux_2}, since $\metric(\mathcal{A}^\star_2|\calV\setminus \calA^\star)=\eta \metric(\mathcal{V}\setminus\calL)$, and $\metric(\mathcal{L}|\mathcal{V}\setminus\calL)-\metric(\mathcal{A}^\star_1|\mathcal{V}\setminus\mathcal{A}^\star_1)\geq 0$;
the latter is true due to the following two observations:~$\metric(\mathcal{L}|\mathcal{V}\setminus\calL)\geq \metric(\mathcal{A}_1^\star|\mathcal{V}\setminus\calL)$, since $\metric$ is monotone and $\mathcal{A}_1^\star \subseteq \mathcal{L}$; and~$\metric(\mathcal{A}_1^\star|\mathcal{V}\setminus\calL)\geq \metric(\mathcal{A}^\star_1|\mathcal{V}\setminus\mathcal{A}^\star_1)$, since $\metric$ is submodular and $\mathcal{V}\setminus\calL\subseteq \mathcal{V}\setminus\mathcal{A}^\star_1$ (see also Fig.~\ref{fig:venn_diagram_for_proof}).

\paragraph{Proof of ineq.~\eqref{ineq:aux_3}} 
To prove ineq.~\eqref{ineq:aux_3}, since $\calA^\star_2\neq \emptyset$ (and, as a result, also $\calL^+\neq \emptyset$), and for all elements $a \in \calL^+$ and all elements $b\in \calA^\star_2$, $\metric(a)\geq \metric(b)$,  from Lemma~\ref{lem:D3} we have:
\begin{align}
\metric(\calA^\star_2|\calL^+)&\leq |\calA^\star_2|\metric(\calL^+)\nonumber\\
&\leq \attack \metric(\calL^+),\label{aux:111}
\end{align}
since $|\calA^\star_2|\leq \attack$.  Overall,
\begin{align}
\metric(\calL^+)&\geq \frac{1}{\attack} \metric(\calA^\star_2|\calL^+)\label{aux5:1}\\
&\geq \frac{1}{\attack}\metric(\calA^\star_2|\calL^+\cup (\calV\setminus \calL)^+)\label{aux5:2}
\end{align}

\begin{align}
&=\frac{1}{\attack}\metric(\mathcal{A}_2^\star|\calV\setminus \calA^\star)\label{aux5:3}\\
&=\eta\frac{1}{\attack}\metric(\calV\setminus \calL),\label{aux5:4}
\end{align}
where ineq.~\eqref{aux5:1} to eq.~\eqref{aux5:4} hold for the following reasons: ineq.~\eqref{aux5:1} follows from ineq.~\eqref{aux:111}; ineq.~\eqref{aux5:2} holds since $\metric$ is submodular and $\calL^+\subseteq \calL^+\cup (\calV\setminus \calL)^+$; eq.~\eqref{aux5:3} holds due to the definitions of the sets $\calL^+$, $(\calV\setminus \calL)^+$ and $\mathcal{A}^\star$; finally, eq.~\eqref{aux5:4} holds due to the definition of $\eta$.  {Overall}, the~latter derivation concludes the proof of ineq.~\eqref{ineq:aux_3}.

\paragraph{Proof of ineq.~\eqref{ineq:aux_4}}  Let $b=1/\attack$.  We complete the proof first for the case where 
$(1-\eta)\geq \eta b$, and then for the case $(1-\eta)<\eta b$: i) When $(1-\eta)\geq \eta b$, $\max\{(1-\eta),\eta b\}= 1-\eta$ and $\eta \leq 1/(1+b)$.  Due to the latter, $1-\eta \geq b/(1+b)=1/(\attack+1)$ and, as a result,~\eqref{ineq:aux_4} holds. ii) When $(1-\eta)< \eta b$, $\max\{(1-\eta),\eta b\}= \eta b$ and $\eta > 1/(1+b)$. Due to the latter, $\eta b >  b/(1+b)$ and, as a result,~\eqref{ineq:aux_4} holds. 

We completed the proof of~$0\leq \eta\leq 1$, and of ineqs.~\eqref{ineq:aux_2},~\eqref{ineq:aux_3} and~\eqref{ineq:aux_4}.  Thus, we also completed the proof of the part $1/(1+\alpha)$ of the bound in the right-hand-side of ineq.~\eqref{ineq:bound_sub}, and, in sum, the proof of ineq.~\eqref{ineq:bound_sub}.

\paragraph*{Proof of ineq.~\eqref{ineq:bound_non_sub}} 
Consider that the objective function $\metric$ is non-decreasing in the active robot set, such that (without loss of generality) $\metric$ is non-negative and $\metric(\emptyset)=0$.

The proof follows the steps of the proof of~\cite[Theorem~1]{tzoumas2018resilientSequential}, by making the following observations:
\begin{align}
& \metric(\mathcal{V}\setminus\mathcal{A}^\star) \nonumber\\
&\;\;\;= \metric(\calL^+\cup (\calV\setminus\calL)^+)\label{ineq2:aux_14}\\
&\;\;\;\geq (1-c_\metric)\sum_{v \in \calL^+\cup (\calV\setminus\calL)^+}\metric(v)\label{ineq2:aux_15}\\
&\;\;\;\geq (1-c_\metric)\left(\sum_{v \in (\mathcal{V}\setminus \mathcal{L})\setminus (\calV\setminus\calL)^+}\metric(v)+\sum_{v \in (\calV\setminus\calL)^+}\metric(v)\right)\label{ineq2:aux_16}\\
&\;\;\;\geq (1-c_\metric)^2\metric\{[(\mathcal{V}\setminus \mathcal{L})\setminus (\calV\setminus\calL)^+]\cup (\calV\setminus\calL)^+\}\label{ineq2:aux_17}\\
&\;\;\;= (1-c_\metric)^2\metric(\calV\setminus\calL)\label{ineq2:aux_18},
\end{align}
where eq.~\eqref{ineq2:aux_14} to~\eqref{ineq2:aux_18} hold for the following reasons: eq.~\eqref{ineq2:aux_14} follows from the definitions of the sets~$\mathcal{L}^+$ and $(\calV\setminus\calL)^+$; ineq.~\eqref{ineq2:aux_15} follows from ineq.~\eqref{ineq2:aux_14} due to Lemma~\ref{lem:curvature}; ineq.~\eqref{ineq2:aux_16} follows from ineq.~\eqref{ineq2:aux_15} because for all elements $v \in \mathcal{L}^+$ and all elements  $v' \in (\mathcal{V}\setminus \mathcal{L})\setminus (\calV\setminus\calL)^+$ we have $\metric(v)\geq \metric(v')$ (note that due to the definitions of the sets~$\mathcal{L}^+$ and $(\calV\setminus\calL)^+$ it is $|\mathcal{L}^+|=|(\mathcal{V}\setminus \mathcal{L})\setminus (\calV\setminus\calL)^+|$, that is, the number of non-removed elements in $\calL$ is equal to the number of removed elements in $\calV\setminus \calL$);  finally, ineq.~\eqref{ineq2:aux_17} follows from ineq.~\eqref{ineq2:aux_16} because the set function $\metric$ is non-decreasing and Corollary~\ref{cor:ineq_from_lemmata} applies.  Overall, the combination of ineq.~\eqref{ineq2:aux_18}  and ineq.~\eqref{ineq:main_ineq} (observe that ineq.~\eqref{ineq:main_ineq} still holds if the objective function $\metric$ is merely non-decreasing) proves ineq.~\eqref{ineq:bound_non_sub}. \hfill $\blacksquare$

\subsection{Proof of Theorem~\ref{th:per_alg_dec_resil_coord_decent}'s Part 2 (Running Time)}\label{proof:runtime-theorem}

\algrobust's running time is found by adding the running time of (i) lines~1-3, \ie $|\calV|\rho$, (ii) line~4, \ie $|\calV|\log(|\calV|)$ (using, \eg the merge
sort algorithm), (iii) lines~5-7, whose running time can be ignored since the optimization problems in line~6 have already been solved in lines~1-3, and (iv) line~8, \ie $\rho$.  The total is  $|\calV|(\rho+1)+|\calV|\log(|\calV|)=O(|\calV|\rho)$. \hfill $\blacksquare$
 
\section{Proof of Proposition~\ref{prop:alg_per_with_coordinate_descent}}\label{app:proof_proposition}

We first prove Proposition~\ref{prop:alg_per_with_coordinate_descent}'s part 1 (approximation bounds), and then, Proposition~\ref{prop:alg_per_with_coordinate_descent}'s part 2 (running time).

\subsection{Proof of Proposition~\ref{prop:alg_per_with_coordinate_descent}'s Part 1 (Approximation Bounds)}

The proof follows the steps of the proof of~Theorem~\ref{th:per_alg_dec_resil_coord_decent}; hence, we describe here only the steps where the proof differs.

\medskip

We first prove ineq.~\eqref{ineq:bound_sub_cd}; then, we prove ineq.~\eqref{ineq:bound_non_sub_cd}.

\paragraph*{Proof of ineq.~\eqref{ineq:bound_sub_cd}}
Consider that the objective function $\metric$ is non-decreasing and submodular in the active robot set, such that (without loss of generality) $\metric$ is non-negative and $\metric(\emptyset)=0$.
Since, per Proposition~\ref{prop:alg_per_with_coordinate_descent}, Algorithm~\ref{alg:dec_resil_coord_decent} calls the coordinate descent algorithm in step~4,  the equivalence in eq.~\eqref{ineq:main_ineq_aux1} is now invalid, and, in particular, using Lemma~\ref{lem:generalized_cd}, the following inequality holds instead:
\begin{equation}\label{ineq:opt_guar_aux}
\metric(\calV\setminus \calL)\geq\frac{1}{2}\max_{\scriptsize\begin{array}{c}
\bar{u}_{i,t} \in \mathcal{U}_{i,t},i\in\calV,\\
 t'=t+1:t+\planninghorizon
\end{array}}\metric(\bar{u}_{\calV\setminus\calL, t+1:t+\planninghorizon}).
\end{equation}
Using ineq.~\eqref{ineq:opt_guar_aux}, and following the same steps as in eqs.~\eqref{ineq:main_ineq_aux1}-\eqref{ineq:main_ineq_aux4}, we conclude:
\begin{equation}
\metric(\calV\setminus \calL)\geq \frac{1}{2} \metric^\star\!\!.\label{ineq4:main_ineq_aux4}
\end{equation}
Using ineq.~\eqref{ineq4:main_ineq_aux4} the same way that ineq.~\eqref{ineq:main_ineq} was used in the proof of Theorem~\ref{th:per_alg_dec_resil_coord_decent}'s part 1, ineq.~\eqref{ineq:bound_non_sub_cd} is proved. 

\paragraph*{Proof of ineq.~\eqref{ineq:bound_non_sub_cd}} 
Consider that the objective function $\metric$ is non-decreasing in the active robot set, such that (without loss of generality) $\metric$ is non-negative and $\metric(\emptyset)=0$. Similarly with the observations we made in the proof of ineq.~\eqref{ineq:bound_sub_cd}, since, per Proposition~\ref{prop:alg_per_with_coordinate_descent}, Algorithm~\ref{alg:dec_resil_coord_decent} calls the coordinate descent algorithm in step~4,  the equivalence in eq.~\eqref{ineq:main_ineq_aux1} is now invalid, and, in particular, using Lemma~\ref{lem:generalized_cd}, the following inequality holds instead:
\begin{equation}\label{ineq:opt_guar_aux_total}
\metric(\calV\setminus \calL)\geq\frac{1-c_\metric}{2}\max_{\scriptsize\begin{array}{c}
\bar{u}_{i,t} \in \mathcal{U}_{i,t},i\in\calV,\\
 t'=t+1:t+\planninghorizon
\end{array}}\metric(\bar{u}_{\calV\setminus\calL, t+1:t+\planninghorizon}).
\end{equation}
Using ineq.~\eqref{ineq:opt_guar_aux_total}, and following the same steps as in eqs.~\eqref{ineq:main_ineq_aux1}-\eqref{ineq:main_ineq_aux4}, we conclude:
\begin{equation}
\metric(\calV\setminus \calL)\geq \frac{1-c_\metric}{2} \metric^\star\!\!.\label{ineq5:main_ineq_aux4}
\end{equation}
Using ineq.~\eqref{ineq5:main_ineq_aux4} the same way that ineq.~\eqref{ineq:main_ineq} was used in the proof of Theorem~\ref{th:per_alg_dec_resil_coord_decent}'s part 1, ineq.~\eqref{ineq:bound_non_sub_cd} is proved. 
\hfill $\blacksquare$

\subsection{Proof of Proposition~\ref{prop:alg_per_with_coordinate_descent}'s Part 2 (Communication Rounds)}

Per \ref{app:description_coordinate}, coordinate descent needs to solve $|\calV|$ times an optimization problem of same complexity as line~2 of \algrobust. \algrobust, instead, needs to solve $|\calV|$ times the optimization problem in line~2 (per the ``for'' loop in lines~1-3), and the optimization problem in line~8, using coordinate descent.  Since the running time of line~4 and lines~5-7 is negligible (cf.~\ref{proof:runtime-theorem}) and can be ignored, the total running time of \algrobust is at most $2\rho_{\scenario{CD}} = O(\rho_{\scenario{CD}})$.  \hfill $\blacksquare$
 
\bibliographystyle{IEEEtran}
\bibliography{references}
 
\begin{biography}[{\includegraphics[width=1in,height=1.25in,keepaspectratio]{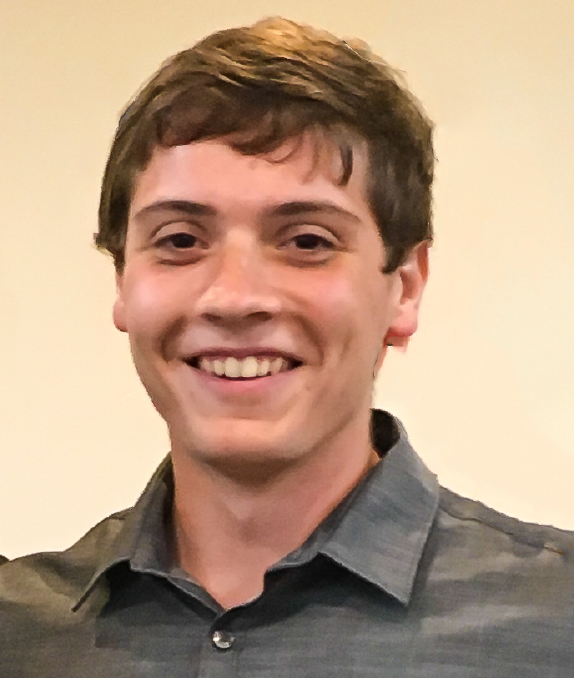}}]{Brent Schlotfeldt} holds a Master of Science in Robotics from the University of Pennsylvania, Philadelphia, PA (2017), and a Bachelor of Science in Electrical Engineering and Computer Science from the University of Maryland, College Park, MD, USA (2016).  Currently, Brent is working towards his Ph.D. in Electrical and Systems Engineering at the University of Pennsylvania, Philadelphia, PA.

His research interests include planning, control, and state estimation for robotic systems, with applications to autonomous driving and active sensing.
\end{biography}

\begin{biography}[{\includegraphics[width=1in,height=1.25in,keepaspectratio]{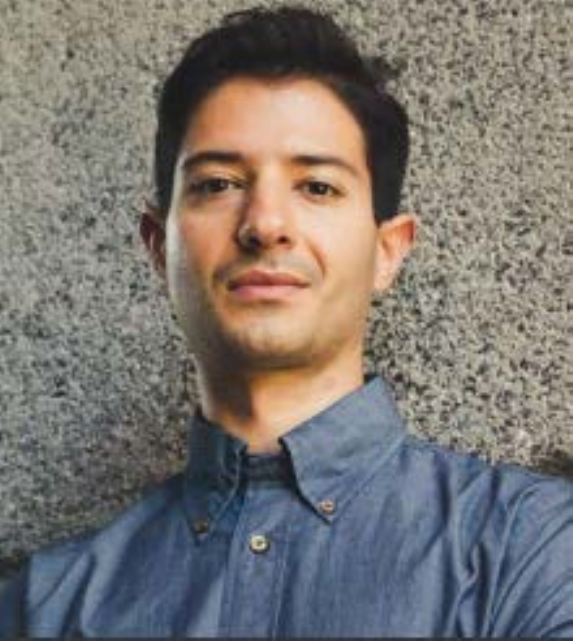}}]{Vasileios Tzoumas} received his Ph.D. in Electrical and Systems Engineering at the University of Pennsylvania (2018). He holds a Master of Arts in Statistics from the Wharton School of Business at the University of Pennsylvania (2016); a Master of Science in Electrical Engineering from the University of Pennsylvania (2016); and a diploma in Electrical and Computer Engineering from the National Technical University of Athens (2012).
	Vasileios is as an Assistant Professor in the Department of Aerospace Engineering, University of Michigan, Ann Arbor. 
	Previously, he was at the Massachusetts Institute of Technology (MIT), in the Department of Aeronautics and Astronautics, and in the Laboratory for Information and Decision Systems (LIDS), were he was a research scientist (2019-2020), and a post-doctoral associate (2018-2019). Vasileios was a visiting Ph.D. student at the Institute for Data, Systems, and Society (IDSS) at MIT during 2017.
	Vasileios works on control, learning, and perception, as well as combinatorial and distributed optimization, with applications to robotics, cyber-physical systems, and self-reconfigurable aerospace systems.  He aims for trustworthy collaborative autonomy.  His work includes foundational results on robust and adaptive combinatorial optimization, with applications to multi-robot information gathering for resiliency against robot failures and adversarial removals.
	Vasileios is a recipient of the Best Paper Award in Robot Vision at the 2020 IEEE International Conference on Robotics and Automation (ICRA), and was a Best Student Paper Award Finalist at the 2017 IEEE Conference in Decision and Control (CDC). 
\end{biography}

\begin{biography}[{\includegraphics[width=1in,height=1.25in,clip,keepaspectratio]{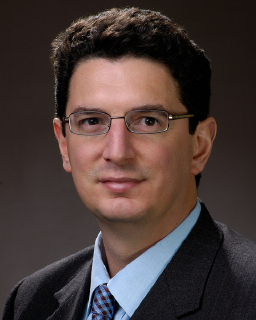}}]{George J.~Pappas} (S'90-M'91-SM'04-F'09) received the Ph.D.~degree in electrical engineering and computer sciences from the University of California, Berkeley, CA, USA, in  1998. He is currently the Joseph Moore Professor and Chair of the Department of Electrical and Systems Engineering, University of Pennsylvania, Philadelphia, PA, USA. He  also holds a secondary appointment with the Department of Computer and Information Sciences and the Department of Mechanical Engineering and Applied Mechanics. He is a Member of the GRASP Lab and the PRECISE Center. He had previously served as the Deputy Dean for Research with the School of Engineering and Applied Science. His research interests include control theory and, in particular, hybrid systems, embedded systems, cyberphysical systems, and hierarchical and distributed control systems, with applications to unmanned aerial vehicles, distributed robotics, green buildings, and biomolecular networks. Dr. Pappas has received various awards, such as the Antonio Ruberti Young Researcher Prize, the George S. Axelby Award, the Hugo Schuck Best Paper Award, the George H. Heilmeier Award, the National Science Foundation PECASE award and numerous best student papers awards.
\end{biography}

\end{document}